\documentclass[pdflatex,sn-mathphys-num]{sn-jnl}% Math and Physical Sciences Numbered Reference Style
\usepackage{graphicx}%
\usepackage{multirow}%
\usepackage{amsmath,amssymb,amsfonts}%
\usepackage{amsthm}%
\usepackage{mathrsfs}%
\usepackage[title]{appendix}%
\usepackage{xcolor}%
\usepackage{textcomp}%
\usepackage{manyfoot}%
\usepackage{booktabs}%
\usepackage{algorithm}%
\usepackage{algorithmicx}%
\usepackage{algpseudocode}%
\usepackage{listings}%
\usepackage{comment}

\usepackage{graphicx}
\usepackage{wrapfig}
\usepackage{lscape}
\usepackage{rotating}
\usepackage{epstopdf}

\usepackage[parfill]{parskip}

\usepackage{graphicx}

\usepackage{float} % make figure appear here
\usepackage{epstopdf}

\usepackage{array}
\usepackage{colortbl}
\usepackage{xcolor}
\usepackage{booktabs}
\usepackage{pdflscape} % For landscape orientation
\usepackage{adjustbox} % For centering table in landscape
\usepackage{afterpage}
\usepackage{placeins}

\usepackage{xcolor}

\usepackage{hyperref}

\usepackage{makecell}

\usepackage{tcolorbox}

\usepackage{array} % for extended table column formatting
\usepackage{colortbl} % for coloured table cells
\usepackage{pdflscape} % for landscape orientation
\usepackage{adjustbox} % for centering tables in landscape
\usepackage{afterpage} % for multi-page figures
\usepackage{placeins} % for \FloatBarrier
\usepackage{hyperref} % for links
\usepackage{makecell} % for multiline cells
\usepackage{tcolorbox} % for boxes
\usepackage{subcaption} % for subfigures
\usepackage{hhline} % for combined table cells
\usepackage[parfill]{parskip}

\theoremstyle{thmstyleone}%
\theoremstyle{thmstyletwo}%

\theoremstyle{thmstylethree}%

\definecolor{headerblue}{rgb}{0, 0.4, 0.8}
\definecolor{red}{rgb}{0.84, 0, 0.25}
\definecolor{orange}{rgb}{0.9, 0.6, 0}
\definecolor{greentext}{rgb}{0, 0.5, 0}
\definecolor{bluehighlight}{rgb}{0.6, 0.8, 1}
\definecolor{grayrow}{gray}{0.95}
\definecolor{purple}{rgb}{0.36, 0.25, 0.83}
\definecolor{turquoise}{rgb}{0.25, 0.71, 0.68}
\definecolor{myblue}{rgb}{0, 0.28, 0.67}
\definecolor{myorange}{rgb}{1, 0.84, 0.50}
\definecolor{myblue2}{rgb}{0.29, 0.56, 0.87}

\begin{document}

\title[Article Title]{A Semi-Automated Annotation Workflow for Paediatric Histopathology Reports Using Small Language Models}


\author*[1,2,3,4]{\fnm{Avish} \sur{Vijayaraghavan}}\email{avish.vijayaraghavan17@imperial.ac.uk}

\author[4]{\fnm{Jaskaran Singh} \sur{Kawatra}}

\author[4]{\fnm{Sebin} \sur{Sabu}}

\author[4]{\fnm{Jonny} \sur{Sheldon}}

\author[3]{\fnm{Will} \sur{Poulett}}

\author[4]{\fnm{Alex} \sur{Eze}}

\author[4]{\fnm{Daniel} \sur{Key}}

\author[4]{\fnm{John} \sur{Booth}}

\author[4]{\fnm{Shiren} \sur{Patel}}

\author[3]{\fnm{Jonny} \sur{Pearson}}

\author[3,4]{\fnm{Dan} \sur{Schofield}}

\author[3]{\fnm{Jonathan} \sur{Hope}}

\author[4]{\fnm{Pavithra} \sur{Rajendran}}

\author[4,5]{\fnm{Neil} \sur{Sebire}}


\affil[1]{\orgdiv{Division of Systems Medicine, Department of Metabolism, Digestion and Reproduction}, \orgname{Imperial College London}, \orgaddress{\city{London}, \country{UK}}}

\affil[2]{\orgdiv{Department of Computing}, \orgname{Imperial College London}, \orgaddress{\city{London}, \country{UK}}}

\affil[3]{\orgdiv{Data Science Team}, \orgname{NHS England}, \orgaddress{\city{Leeds \& London}, \country{UK}}}

\affil[4]{\orgdiv{DRE Team, GOSH DRIVE}, \orgname{Great Ormond Street Hospital}, \orgaddress{\city{London}, \country{UK}}}

\affil[5]{\orgdiv{Population, Policy \& Practice Department}, \orgname{University College London}, \orgaddress{\city{London}, \country{UK}}}


\abstract{
Electronic Patient Record (EPR) systems contain valuable clinical information, but much of it is trapped in unstructured text, limiting its use for research and decision-making. Large language models can extract such information but require substantial computational resources to run locally, and sending sensitive clinical data to cloud-based services, even when deidentified, raises significant patient privacy concerns. In this study, we develop a resource-efficient semi-automated annotation workflow using small language models (SLMs) to extract structured information from unstructured EPR data, focusing on paediatric histopathology reports. As a proof-of-concept, we apply the workflow to paediatric renal biopsy reports, a domain chosen for its constrained diagnostic scope and well-defined underlying biology. We develop the workflow iteratively with clinical oversight across three meetings, manually annotating 400 reports from a dataset of 2,111 at Great Ormond Street Hospital as a gold standard, while developing an automated information extraction approach using SLMs. We frame extraction as a Question-Answering task grounded by clinician-guided entity guidelines and few-shot examples, evaluating five instruction-tuned SLMs with a disagreement modelling framework to prioritise reports for clinical review. Gemma 2 2B achieves the highest accuracy at 84.3\%, outperforming off-the-shelf models including spaCy (74.3\%), BioBERT-SQuAD (62.3\%), RoBERTa-SQuAD (59.7\%), and GLiNER (60.2\%). Entity guidelines improved performance by 7-19\% over the zero-shot baseline, and few-shot examples by 6-38\%, though their benefits do not compound when combined. These results demonstrate that SLMs can extract structured information from specialised clinical domains on CPU-only infrastructure with minimal clinician involvement. Our code is available at \url{https://github.com/gosh-dre/nlp_renal_biopsy}.}

\keywords{small language models, information extraction, named entity recognition, question-answering, medical reports, paediatrics, renal biopsy}



\maketitle

\section{Introduction}

Data-driven clinical decision-making enables healthcare professionals to deliver personalised care by drawing on the diverse digitised information stored within Electronic Patient Record (EPR) systems \cite{lyu2025data}. However, much of this pertinent information remains locked within unstructured data. Without conversion to a standardised structured format, it cannot be used easily for quantitative analysis or secondary research.

\begin{figure}[h]
    \centering
    \includegraphics[width=1.\linewidth]{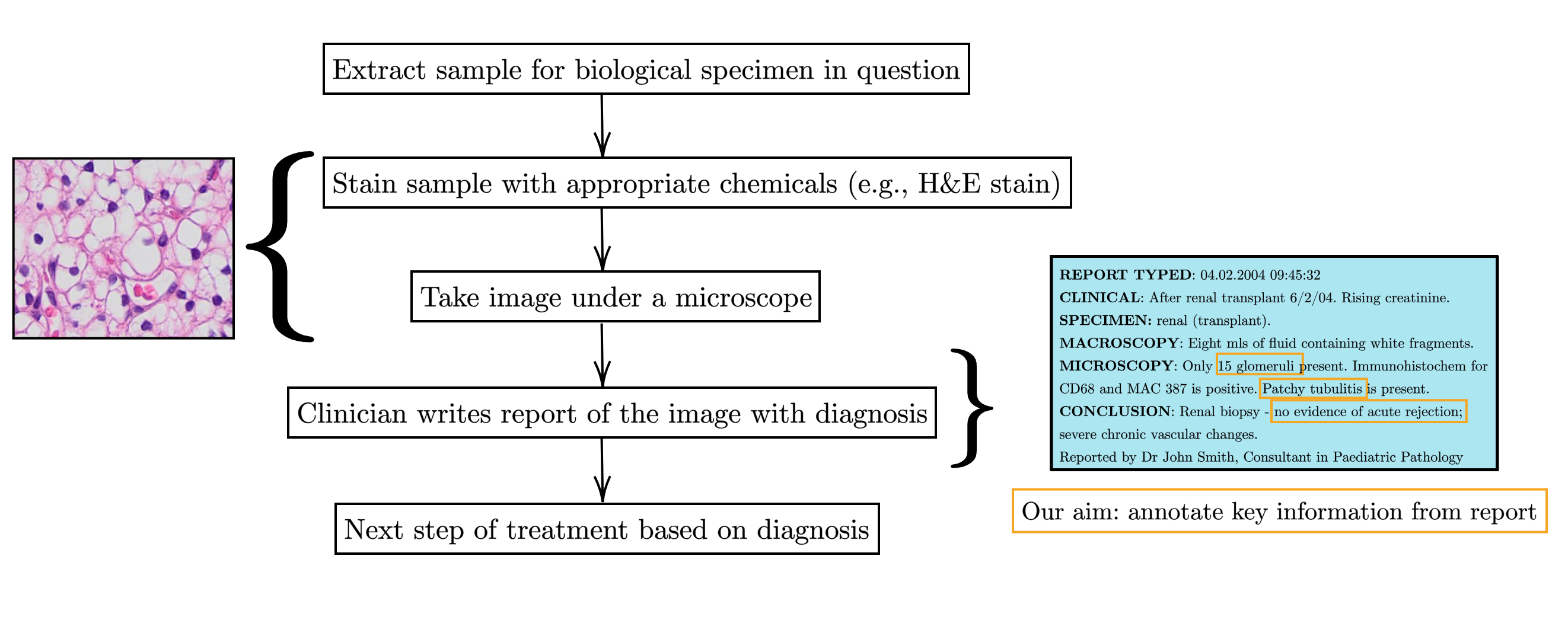}
    \caption{The general histopathology workflow. Our focus is on extracting key text spans from the clinician's report that are relevant for diagnosis.}
    \label{fig:histo}
\end{figure}

Histopathology, the microscopic examination of tissue samples, is fundamental to disease diagnosis and treatment planning. A simplified process is shown in Figure \ref{fig:histo}: a biopsy is performed, tissue is stained, pathologists examine the slides and document their findings in detailed reports that inform clinical decision-making. However, these reports exemplify the challenge of unstructured clinical text, combining clinical history, detailed microscopic and macroscopic observations, and diagnostic conclusions in free-form narratives that make it difficult to aggregate findings. Automating the annotation of histopathology reports could enhance patient care in multiple ways: standardising the documentation process, reducing clinician workload, enabling scalable insights for research, and supporting quality audits. 

A fundamental challenge in working with histopathology reports lies in the nature of medical free text. These reports contain domain-specific terminology, abbreviations, informal language, and occasionally noisy or misspelled text. A comprehensive study by the King's Fund Trust \cite{bean2023hospital}, analysing over one million clinical documents, confirmed that a significant portion of clinical communication occurs in free text format, highlighting the importance of addressing this domain-specific language variability. Additionally, these reports often include ambiguities and shorthand notations that, while intuitive for human clinicians, pose significant challenges for traditional natural language processing (NLP) models.

Dictionary or ontology-based methods (e.g., SNOMED-CT) work well for simple diagnosis prediction, but our focus is on capturing detailed contextual information for accurate report summarisation. Token-based named entity recognition (NER) methods can achieve high performance on clearly-defined entities, but struggle with variable phrasing, longer and often discontinuous text spans, and implicit entity relations. Additionally, histopathology reports often contain meta-annotations like negations (e.g., ``there is no evidence of rejection") and uncertainties (e.g., ``there are features suggestive of chronic tubulointerstitial nephritis"). Standard medical NER models like medspaCy \cite{eyre2022launching} do not support meta-annotations, and while MedCAT \cite{kraljevic2021multi} does, neither adequately captures the variability of entity descriptions found in histopathology reports.

The emergence of large language models (LLMs) presents an opportunity to address these challenges. Unlike token-classification-based methods, autoregressive LLMs perform well on language tasks that do not require precise localisation or categorisation of entities, making them better suited to capturing variable and context-dependent information. LLMs also offer greater task flexibility because their inputs (``prompts") can be customised in natural language and easily adapted to specific biomedical domains. This flexibility makes them straightforward to integrate into existing clinical workflows. However, most approaches require substantial computational resources and external API access, making them unsuitable for sensitive paediatric data that must remain on local infrastructure.

In this study, we propose an automated workflow for annotating unstructured text within medical reports using a Question-Answering (QA) framework. We describe its development using real-world data from Great Ormond Street Hospital, a context that introduces several constraints: (1) the sensitive nature of the data restricts access to secure on-premises servers, (2) computational infrastructure is limited to CPU processing only, (3) clinician time for annotation tasks is scarce, and (4) large volumes of data remain unlabelled. These challenges are compounded by the inherent complexity of paediatric medicine, where patient presentations vary considerably, and by the hospital's national referral pathways, which bring patients from diverse geographic and environmental backgrounds that may influence disease presentation. This is particularly relevant given that few language models have been trained on or applied to paediatric text \cite{jabarulla2024meddoc, barile2024diagnostic, andrew2024evaluating, yang2024pediatricsgpt}, despite - or perhaps due to - the known high variability in paediatric clinical reporting.

In order to overcome these constraints, our approach is focussed on using LLMs in the order of 1B to 5B parameters, also referred to as \textbf{Small Language Models} (SLMs). To address the scarcity of labelled data, our workflow includes a Streamlit-based annotation interface for efficient labelling. A subsequent disagreement modelling stage prioritises data scientist-annotated reports for secondary clinical review. We have investigated different quantised generative models and evaluated their performance using modified prompts for in-context learning. This approach offers a pragmatic balance between computational efficiency and the ability to capture complex medical narratives, alongside easier interfacing with non-computational practitioners. 

We demonstrate our approach on renal biopsies, a relatively constrained and well-understood domain within histopathology. Renal biopsies involve a narrower range of pathological triggers and diagnostic outcomes, making them ideal for testing the robustness of our proposed machine learning pipeline alongside human evaluation. Since histopathology as a whole is a well-established diagnostic modality, renal biopsy reports offer a reliable and informative starting point, acting as a proof-of-concept for more complex and less standardised pathology subfields.

\section{Related Work}

\subsection{LLMs for Biomedical NER into QA}

The emergence of word embeddings marked a shift towards deep learning-based biomedical NER. scispaCy \cite{neumann2019scispacy}, for instance, integrated UMLS linkages and offered pre-trained embeddings tailored to biomedical texts. Biomedical takes on the transformer architecture BERT (e.g., BioBERT \cite{lee2019biobert}, ClinicalBERT \cite{alsentzer2019clinicalbert}), significantly advanced the field by capturing contextual information and long-range dependencies. MedCAT \cite{kraljevic2021multi} built upon this by incorporating spelling correction, contextual disambiguation (e.g., differentiating ``hr'' as ``hour'' or ``heart rate''), and flexible entity linking to UMLS and SNOMED-CT, improving standardisation potential. These BERT-based models outperform earlier systems on benchmark datasets but often require fine-tuning on large annotated corpora, which limits their applicability in resource-constrained environments.

Recent NER models have focussed on adaptability to unseen entities at test time, often using BERT backbones within their overall architectures. GLiNER \cite{zaratiana2024gliner} approaches NER as a compatibility task between entity type prompts and text spans, using two-layer feedforward neural networks to compute embeddings for both components. The model uses a cross product of these embeddings wrapped in a sigmoid to calculate the probability that a span $(i,j)$ is of entity type $t$. GLiNER calculates match probabilities through cross-product scoring with sigmoid activation, making it flexible enough to support user-defined labels and handle flat and fully-nested NER. However, GLiNER cannot process discontinuous spans or partially-nested NER, limiting its application in medical reports where more complex expressions of symptoms are possible. 

Question answering (QA) frames information extraction as a reading comprehension task, where models must locate and extract relevant information from input text to answer specific questions. While NER outputs are limited to entity-type pairs, QA can generate free-form answers that combine multiple pieces of information, reason across paragraphs, and handle implicit relationships between concepts. There is some form of information synthesis that happens in QA unlike NER. Alongside BERT-backbone QA models, the field has also explored using generative LLMs (i.e., unidirectional architectures building off GPT models, unlike the bidirectional BERT-based architectures that predict masked tokens) for NER by reformulating NER as a text generation task \cite{wang2025gptner, ashok2023promptner}. This approach includes input sentences in pre-defined prompt templates, treating entity annotation as a fill-in-the-blank (i.e., text generation) problem as opposed to traditional sequence labelling. More recent methods like NuNER \cite{bogdanov2024nuner} and NuExtract \cite{numind2024nuextract} have extended these ideas by combining BERT-based models with data generated from GPT-style models and fine-tuning generative models for structured output extraction, respectively. These prompt-based NER models exist in the hazy space between NER and QA, or to be even more precise, the hazy space between extractive QA (where answers must be in the text) and open generative QA (where answers can be generated based on the text). QA is useful because it can capture document-level context but is not without flaws - it is more computationally expensive, and, when using generative inference techniques, prone to making things up \cite{huang2025hallucination}.

\subsection{LLMs for Pathology Reports}

Large language models (LLMs) have been applied to a variety of pathology reports, with different focuses: domain-specific tokenisation, active learning, data augmentation, large datasets, and localised models. We note that the majority of these papers have much larger datasets than our dataset of 2,462 reports, bar Zeng et al's paper \cite{zeng2023improving}, and our focus is mostly on human-integrated annotation workflows to maximise reliability and interpretability.

PathologyBERT \cite{santos2023pathologybert} introduced a custom tokeniser to address limitations of the WordPiece tokeniser, which can break down medical terms in ways that lose semantic meaning (e.g., ``carcinoma" into [``car'', ``cin'', ``oma'']). Using 340,492 unstructured histopathology reports from 67,136 patients at Emory University Hospital between 1981-2021 alongside with labelled set of 6,681 reports from 3,155 patients, they demonstrated greater coverage of pathology-specific terminology compared to previous clinical language models like BlueBERT \cite{peng2020empirical} and ClinicalBERT \cite{alsentzer2019clinicalbert}. Their reports followed a semi-structured template like ours: history, diagnosis, gross description, microscopy examination, and comments. However, unlike our data, their reports are much shorter with an average report length of 42 $\pm$ 26 tokens. 

Mu et al. \cite{mu2021bert} generated diagnostically relevant embeddings from 11,000 bone marrow pathology synopses using a BERT model enhanced with active learning. The synopses followed the structure of: report (tissue morphology), diagnosis (information on the pathology specimen, ancillary testing, clinical history). The active learning approach minimised manual annotation efforts while achieving strong performance, with only 350 annotated synopses required to reach an F1 score plateau, outperforming random sampling strategies. Zeng and colleagues \cite{zeng2023improving} trained a BERT-based NER model on 1,438 annotated US pathology reports and tested it on a separate dataset of 55 reports from the UAE. They found that data augmentation strategies, such as mention replacement, synonym replacement, label-wise token replacement, and segment shuffling, led to more accurate entity recognition for cancer grades, subtypes, and lesion positions compared to LSTM models. 

Lu et al. \cite{lu2023assessing} analysed two key datasets, (1) a dataset of 93,039 reports with CPT codes, and (2) a larger general dataset of 749,136 reports, examining various classification tasks. Their analysis revealed differences in entity complexity across categories, with position-related entities showing high variability while cancer grades had more constrained vocabularies. They also found that medical domain pretraining did not consistently improve performance. Bumgardner and colleagues \cite{bumgardner2024local} explored the use of locally-hosted large language models (LLMs) for extracting structured information from surgical pathology reports. The task was complicated by the presence of multiple tissue specimens with varying levels of malignancies within a single case. Their reports followed a rough schema of gross descriptions (characteristics of tissue specimens), final diagnosis (diagnosis based on microscopy, lab results and clinical notes), and condition codes. Using 150,000 reports, they evaluated several models, finding a quantised LLaMA 13B model outperformed others, including the Path-LLaMA variant which was instruction-tuned for pathology tasks.

\subsection{LLMs for Human-Integrated Annotation Workflows}

A recent review paper \cite{tan2024large} has split this nascent field up into three categories: annotation generation, annotation assessment, and annotation utilisation. The paper has further categorisation of eight tasks, of which our work loosely falls into four: categorising raw data with labels, adding intermediate labels for contextual depth, confidence scores for assessing annotation reliability, and tailoring outputs to user needs. The review takes a bigger focus on knowledge distillation - our work differs by emphasising asynchronous human-integrated annotation workflows using smaller models in medical settings. Here, we discuss some relevant example papers and how they have guided our final annotation workflow in Figure \ref{fig:workflow}.

One common approach is to use iterative strategies for annotation. FreeAL demonstrates an iterative knowledge distillation framework where GPT-3.5 Turbo generates weak labels for unsupervised data, while RoBERTa-Base filters for high-quality annotations \cite{xiao2023freeal}. This recursive process addresses LLM confirmation bias through iterative refinement, where the smaller model improves the annotation pool to enhance the LLM's subsequent performance. In the biomedical domain, Munnangi et al. \cite{munnangi2024fly} developed an entity definition-specific approach combining UMLS concept definitions with entity linking . Their two-stage approach for NER allowed models to use definitions to revise predictions through entity addition, removal, or type reassignment, while using scispaCy for mapping biomedical concepts to a human-curated set of relevant UMLS entries. Naraki et al. \cite{naraki2024augmenting} use label mixing to deal with entity class imbalance, prompt templates with strict entity definitions and missing entity rules, and a downstream BERT model to evaluate annotations. This work is the most similar reference point to our study; however, we have fairly balanced entity occurrence, take a broader focus on the workflow, and choose to incorporate expert knowledge specific to medical reports.

A broad philosophy for ensuring reliable annotations without humans is some form of ensembling. The T-SAS framework \cite{jeong2023test} demonstrated effective annotation using unlabelled test data through stochastic answer generation, employing Monte-Carlo dropout with majority voting and ensemble filtering to mitigate noise. A blog post by Warmerdam \cite{warmerdam2023disagreement} recommends disagreement modelling as a simpler alternative to active learning for refining annotations, using error analysis in a feedback loop to produce better inputs for the models (e.g. refining prompts). Stochastic methods with filtering is a useful strategy which we have not been able to use due to compute limitations but served as equal inspiration to using disagreement modelling in our eventual workflow (Figure \ref{fig:workflow}). 

An important consideration for us is finding the critical parts of the annotation workflow where expert knowledge is needed. Explosion (an industry NLP lab) has emphasised finding the balance between humans and LLMs by considering their respective strengths and limitations \cite{montani2024distillation}. Humans are great at contextual understanding and resolving ambiguity but worse than machines at tasks requiring extensive memory and consistent repetition. They note we have strong limitations on how much information we can understand at one time and suggest focussing on the minimum required information for a human when requesting feedback. A recent paper by Gao et al. \cite{gao2024optimising} proposed CliniCoCo - a framework aligning AI models with human-in-the-loop feedback for better clinical coding workflows. CliniCoCo has a range of human touchpoints: selective annotation during preprocessing, active learning during training, and customisable explainable AI workflows at the final clinical decision-making stage. The findings from Explosion and CliniCoCo were inspiration for minimising and optimising the expert touchpoints in our final workflow.

\section{Materials and Methods}

\subsection{Study Design and Setting}
For the study, we collected retrospective paediatric histopathology reports from renal biopsies carried out at Great Ormond Street Hospital from January 2018 to December 2023. The aim of this study was twofold: (1) to evaluate an approach that reduces reliance on clinician annotation while maintaining reliable extraction, and (2) to investigate the ability of techniques that do not require fine-tuning - in particular, prompt engineering with SLMs - to extract complex symptomatic information and final diagnoses from medical reports. Analysis was conducted on a standard GOSH laptop with 16GB RAM (the same as a standard NHS laptop).

Success criteria for the study were: (1) reasonable inference time on the hospital on-premises infrastructure (defined as two days as per the threshold for an ``extended run time" for GOSH laptops \cite{pope2024real}), (2) acceptable accuracy (defined with a clinician as $\geq$80\%), and (3) minimal impact to existing clinician workflows. These are specified in Figure \ref{fig:workflow}.

\subsection{Dataset and Preprocessing}

\textbf{Renal Biopsy Report Dataset.} We downloaded a raw dataset of 2,462 renal biopsy histopathology reports from 1,581 paediatric patients at Great Ormond Street Hospital between January 2018 and December 2023. Renal biopsy pathology interpretation focuses on two key compartments within each kidney's million nephrons: (1) the glomeruli, where plasma filtration occurs with retention of larger molecules like proteins while allowing smaller molecules and water to pass into Bowman's space, and (2) the tubulointerstitial compartment comprising tubules and surrounding interstitial tissue. The pathological findings differ between native and transplant biopsies. Native kidney biopsies evaluate primary glomerular diseases, tubulointerstitial disorders, and vascular pathology, while transplant biopsies primarily assess for rejection, manifested by tubulitis (inflammatory cell invasion of tubular epithelium) and interstitial infiltrates in the tubulointerstitial space. Key pathological changes include \textbf{sclerosis} (scarring within glomerular and vascular structures) and \textbf{fibrosis} (scarring of the tubulointerstitial compartment).

\textbf{Preprocessing and Exploratory Data Analysis}
Reports within the hospital setting followed a standard structure that includes the sections \emph{clinical (history)}, \emph{(biopsy) specimen}, \emph{macroscopy (i.e., gross description)}, \emph{microscopy} and \emph{conclusion}. In this work, we are interested in the extraction of entities corresponding to disease processes and hence our focus is on the text present within the \emph{microscopy} and \emph{conclusion} sections. A rule-based approach is used for extracting the contents corresponding to the different sections. Results of the preprocessing steps showed that 2,111 reports contains the required sections for analysis, exhibiting a mean length of 854 $\pm$ 358 words, ranging from 50 to 3,493 words, with the maximum length corresponding to approximately 4,656 tokens. Word frequency analysis revealed that terms associated with our entities appeared prominently across all reports after removing common stop words, with the \emph{microscopy} section consistently being the longest. Figures for this exploratory data analysis can be viewed in Supplementary Section 2.

A random sample of 400 reports (18.9\% of the final dataset) was manually annotated for ground truth comparison, with 100 of these reports used to establish the entity schema and annotation workflow. To validate the narrowness of the renal biopsy pathway, we performed a qualitative investigation of diagnoses in the selected reports, revealing 13 unique conditions with all rejection grades (none to grade 3) represented, including \emph{chronic vasculopathy}, \emph{BK virus}, \emph{chronic allograft nephropathy}, \emph{tubular necrosis}, \emph{urinary tract infection (UTI)}, \emph{acute pyelonephritis} (a serious type of UTI), \emph{cortical necrosis}, \emph{lupus nephritis}, and varying forms of \emph{glomerulonephritis}. For completeness, we note there were also rarer conditions that only appeared once in our sample: \emph{cystic hamartoma of renal pelvi}, \emph{tubulointerstitial nephritis}, \emph{parenchymal infarction}, and \emph{renal artery thrombosis}.

\subsection{Entity Guidelines and Prompt Creation}
Our entity schema builds off the Banff classification \cite{loupy2022thirty}, a renal transplant staging system, to accommodate both transplant and native (i.e., non-transplant) biopsies. This adaptation uses fewer, more general entities while maintaining detailed classification of glomerular patterns and reference to other blood vessel abnormalities. An initial set of entities were identified with an expert consultant in paediatric pathology. Using this initial schema, we extracted key entities from 100 reports (approximately 4\% of the dataset) over two weeks, analysing their variations and associated meta-annotations such as negation and uncertainty.

We then built a more detailed entity guidelines schema that sought to balance biological validity (entities can comprehensively describe mechanisms underlying kidney conditions in our dataset), clinical utility (entities are relevant for clinical decision-making) and simplicity (a minimal set of entities). We also detailed the possible entity types, namely: (1) binary or boolean (T/F), (2) categorical, (3) numerical, (4) string-simple (a proper noun or a short phase) and (5) string-complex (longer strings, often used for descriptions or explanations). Schema complexity is a function of number of entities and entity type complexity. The simplest case is a minimal set of clinically-actionable binary entity types - this is both easy to evaluate and turn into a subsequent decision - but unfortunately, may lose key information about the pathological process, and lose interpretability if reasoning is required. Final entities with their data types included: \textbf{cortex\_present} (\emph{binary}), \textbf{medulla\_present} (\emph{binary}), \textbf{n\_total\_glomeruli},  (\emph{numerical}; simplified in text to \textit{n\_global}), \textbf{n\_segmental\_sclerosed\_glomeruli} (\emph{numerical}; simplified in text to \textit{n\_segmental}), \textbf{n\_global\_sclerosed\_glomeruli} (\emph{numerical}; simplified in text to \textit{n\_global}), \textbf{abnormal\_glomeruli} (\emph{binary}), \textbf{chronic\_change} (\emph{numerical/string-simple}), \textbf{transplant} (\emph{binary}), and \textbf{final\_diagnosis} (\emph{string-complex}; sometimes simplified in text to \textit{diagnosis}). 

Prompt creation requires creation of a minimal text string that describes the task and includes the report information. Where applicable, related entity types are grouped for simultaneous examination in a prompt question. Building on questions that arose while developing an NER pipeline for Tagalog \cite{miranda2023annotation, miranda2023developing}, this iterative process was guided by four central questions: 
\begin{enumerate}
\item Are these entities relevant to the biology being studied? 
\item Do entities share characteristics with other entities (e.g. contain similar terms/biology, appear in similar region of report)?
\item Is it beneficial for the model to learn particular edge cases and can this be encoded as additional prompt information?
\item Are any guidelines vague or confusing - should we remove or update them?
\end{enumerate}

\subsection{Experimental Settings}
Given the challenges of entity variability and relations alongside the aims of a generalisable pipeline, we decided to take a more flexible model class and frame the task as question answering rather than NER. Given that we have a limited amount of data, we chose to move to modern iterations of pretrained autoregressive models (e.g. BERT, GPT, etc.) which can perform well out-the-box with minimal finetuning or in-context learning.

Given our privacy and infrastructure constraints, we explored SLMs on the order of \textbf{1B}-\textbf{5B} parameters. All models were run locally in keeping with our privacy constraints, with options for two local inferencing servers: \texttt{Ollama} and \texttt{llama-cpp-python}. We used five instruction-tuned models: \textbf{Qwen-2.5 1.5B FP16}, \textbf{Gemma 2 2B FP16}, \textbf{Llama 3.2 3B Q8}, \textbf{Phi-3.5 Mini 3.8B Q4}, \textbf{Phi-3.5 Mini 3.8B Q8}. These models were selected based on our infrastructure requirements: operation within 16GB RAM constraints and reasonable inference times, with preference given to models requiring minimal quantisation.

Experimental settings included:
\begin{enumerate}
    \item Zero-shot SLM prediction without guidelines
    \item Two-shot SLM prediction without guidelines
    \item Zero-shot SLM prediction with guidelines provided
    \item Two-shot SLM prediction with guidelines provided
\end{enumerate}

The two-shots consisted of synthetic exemplar report-annotation pairs of varying report ``complexity" and guidelines corresponded to rules for each entity.
 
The quality of prediction was measured by accuracy, calculated as the proportion of correctly extracted entities across all reports. For entity types described in the guidelines, we used exact string matching to compare binary, categorical, and numerical types, and a smaller 1B parameter LLaMA 3.2 language model (LM) under the LM-as-a-Judge approach to compare whether the string-complex entities match. When using this LM-as-a-Judge method, we asked true/false questions with the temperature set to zero and the maximum output tokens set to two, maximising the chance of a ``True" or ``False" response and preventing subsequent token generation. LM judges are known to exhibit biases including position bias (where models' judgements are influenced by the order in which options are presented) \cite{zheng2023judging} and self-preference bias (where LMs favour text generated by themselves or similar model families over text from other sources) \cite{panickssery2024llm}. We tested for position bias by swapping entity order in a set of 147 example entity pairs and found 93.9\% of cases outputted the same result (the symmetry bar in the final plot of Supplementary Figure 10), indicating minimal, though not insignificant, position bias in our judge model. We acknowledge that our judge model (LLaMA 3.2 1B) shares the same architecture family as one of the evaluated models, presenting a potential source of self-preference bias in our evaluations. All entities are weighted equally for simplicity and so a schema of four entities will have a quarter (1/4) point per entity per report. In practical settings, varied entity weighting might be preferred, but this is sufficient for our proof-of-concept.

\subsection{Alternative Model Comparison}

We compared our approach against several alternative models, transforming all outputs to match our QA format for standardised evaluation. The baseline models included: (1) spaCy with entity extraction rules converted from our guidelines into regex and conditional logic, (2) BERT-based QA models fine-tuned on SQuAD \cite{rajpurkar2018know}, using both a general RoBERTa model \cite{liu2019roberta} and BioBERT model (pre-trained on biomedical text) \cite{lee2019biobert}, (3) GLiNER \cite{zaratiana2024gliner} with a biomedical BERT backbone for named entity recognition, and (4) NuExtract \cite{numind2024nuextract}, a structured extraction model. We provided spaCy rules that were as comprehensive as possible based on our guidelines and reading of the reports, took entity questions from our SLM prompts to use as the questions required for BERT-QA models, and used the entity categories directly with GLiNER and NuExtract.

\subsection{Disagreement Modelling Framework}

We implement a disagreement modelling framework to evaluate the concordance between two machine learning models and ground truth annotations for additional reliability across models and to prioritise certain reports for clinical review. The framework performs three-way comparisons between ground truth labels and predictions from two different models, classifying agreement patterns into five categories: ``All\_Agree\_Correct" (both models predict correctly), ``All\_Disagree" (both models predict incorrectly with different predictions), ``Both\_Models\_Wrong\_Same" (both models predict incorrectly with the same prediction), ``Model1\_Correct\_Model2\_Wrong", and ``Model2\_Correct\_Model1\_Wrong". As with our LM-as-a-Judge function, for complex string-based entities, the framework uses an SLM evaluator to determine equivalence between predictions and ground truth values  (with the same potential biases noted above), while simpler categorical and numerical entities use exact matching. The framework incorporates clinical priority weighting to identify cases requiring urgent review, with higher weighting given to entities with greater diagnostic significance. High priority (weighting of 3) entities are \textit{n\_glomeruli}, \textit{n\_global}, \textit{abnormal\_glomeruli}, \textit{transplant}, and \textit{final\_diagnosis}; medium priority (weighting of 2) are \textit{chronic\_change}, \textit{cortex\_present}, and \textit{medulla\_present}; and the one low priority (weighting of 1) entity is \textit{n\_segmental}. 

\subsection{Code Implementation of Workflow} \label{code-implementation}

We have designed modular software that maps the annotation workflow into three core components, centred around a project-specific guidelines spreadsheet called \texttt{guidelines.xlsx} (shown in Table \ref{tab:entity-guidelines}). This spreadsheet defines medical entities and phrases to be extracted from reports, their codes, annotation guidelines, and resulting prompts in a format accessible to non-computational researchers. The three core components are: (1) \texttt{preprocessor.py} for processing medical reports, (2) \texttt{qa.py} containing the general system message and project-specific question-answering logic for \texttt{Ollama} or \texttt{llama-cpp-python} models, and (3) \texttt{annotation\_app.py}, a Streamlit-based application for displaying relevant report sections during annotation. These components balance standardisation with flexibility with shared base classes for core software and project-specific modules for preprocessing and few-shot examples.

An additional fourth component enables automated annotation through disagreement modelling, where two separate SLM models generate annotated JSON files and flag reports with mismatches above a specified threshold for clinician review. This can also be performed through a notebook if models have already been run and annotations generated. A second Streamlit application, \texttt{comparison\_app.py}, displays predicted results from automated annotation, allowing the clinician validators to comment on mismatched entities. More information on annotation processes and the code can be found in Supplementary Section 1, with the complete implementation available as an open-source Python codebase at \url{www.github.com/gosh-dre/nlp_renal_biopsy}.

\subsection{Patient and Public Involvement}
Given the retrospective nature of this study using existing clinical reports, direct patient involvement in the research design was not feasible. However, our approach was informed by previous GOSH research showing that paediatric patients prefer AI systems that prioritise accuracy over speed and favour clinician-in-the-loop workflows \cite{lee2024would}. The annotation workflow was specifically designed to ensure high performance through use of domain knowledge and clinical oversight, while maintaining efficiency through shifting the workload to data scientists and semi-automated annotation pipelines — aligning with these preferences.

\section{Results}

\subsection{Iterative Annotation Workflow with Expert Touchpoints}

\begin{figure}[h]
    \centering
    \includegraphics[width=0.79\linewidth]{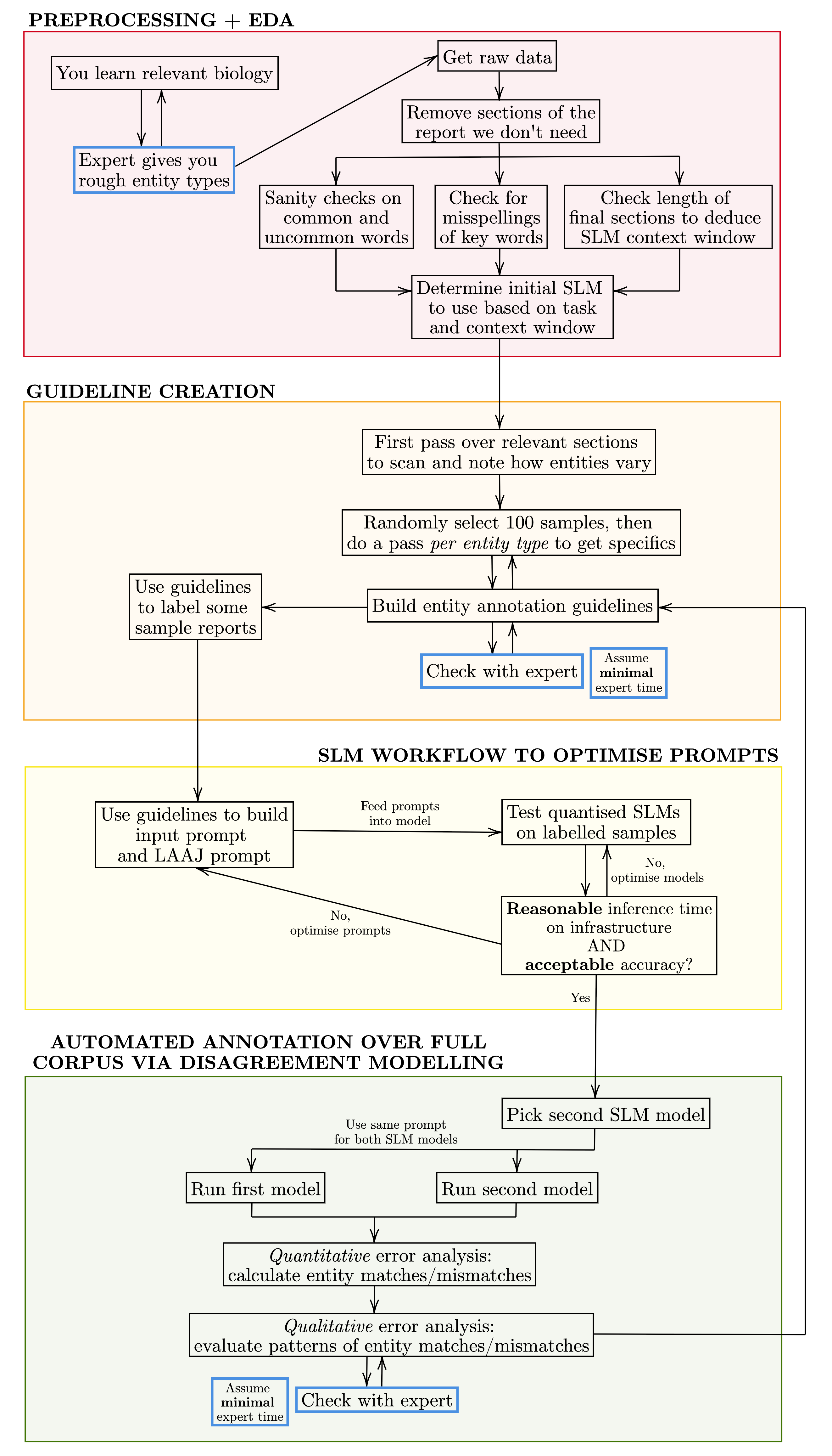}
    \caption{Annotation workflow for medical reports which three expert touchpoints highlighted with \textcolor{myblue2}{\textbf{thick blue lines}}. The second and third touchpoints have potential to be indefinitely long and so we have specified the assumption of minimal expert time.}
    \label{fig:workflow}
\end{figure}

Our overall workflow for annotating reports and performing question answering, shown in Figure \ref{fig:workflow}, was developed as we iterated over the 100 report sample (4.7\% of the final dataset). It consists of four overarching stages in a feedback loop with three key clinician touchpoints: (1) preprocessing the raw data into an appropriate format and performing exploratory data analysis, (2) creation of the annotation guidelines, and (3) an SLM workflow where prompts are created from the guidelines and optimised, before (4) the final automated annotation workflow. We describe the overall workflow as semi-automated to emphasise the importance of data scientist and clinician input throughout.

\subsection{Entity Schema and Guidelines Development}

Qualitative analysis of the first 100 reports (of the 400 we annotated in total) revealed significant entity variability requiring systematic guideline development. We developed five core principles for creating entity guidelines (Table \ref{tab:entity-guidelines}): allowing data type changes, handling missing entity cases, defining biological scope, specifying phrases that confirm/negate the entity, and mapping severity scores. Our final schema evolved from the initial seven entities (Table \ref{tab:initial-entities}) to nine refined entities, including the separation of sclerosed glomeruli into segmental and global categories, and the combination of cortex and medulla presence into a grouped entity due to their frequent co-occurrence.

Key modifications included changing the chronic change entity from purely numerical to numerical/string-simple to accommodate the frequent use of qualitative descriptors (e.g., ``mild", ``moderate", etc.) over precise percentages, and converting the final diagnosis from categorical codes to complex strings to capture the full diagnostic context including rejection grades. The guidelines explicitly addressed edge cases such as implicit information requiring inference (e.g., sclerotic glomeruli count may not be mentioned but should be zero when all abnormalities are absent) and meta-annotations including negations and uncertainties commonly found in pathology reports.

\begin{table}[h]
\centering
\fontsize{12}{14}\selectfont
\caption{Initial entities and their data types from the consultant clinician.}
\begin{tabular}{|l|l|}
\hline
\textbf{Entity Name} & \textbf{Data Type} \\ \hline
cortex & Binary \\ \hline
medulla & Binary \\ \hline
number of glomeruli & Numerical \\ \hline
number of sclerosed glomeruli & Numerical \\ \hline
\% chronic change/fibrosis & Numerical \\ \hline
glomeruli normal or abnormal & Binary \\ \hline
final diagnosis & Categorical/Code \\ \hline
\end{tabular}
\label{tab:initial-entities}
\end{table}

\begin{landscape} % Start landscape mode
\begin{table}[h!]
    \centering
    \vspace{2cm}
    \adjustbox{max width=1.5\textwidth}{
    \fontsize{10}{12}\selectfont
    \begin{tabular}{|p{4cm}|p{4cm}|p{13cm}|}
    \hline
    %\rowcolor{bluehighlight}
    \textbf{Entity type} & \textbf{Data type} & \textbf{Guidelines} \\
    \hline
    cortex / medulla present & \textcolor{orange}{Tuple(Binary, Binary)} &
    \textcolor{red}{- Lack of mention of either implies absence. But if neither mentioned, assume cortex.} \\
    \hline
    number of glomeruli & Numerical &
    \textcolor{purple}{- All we care about is the total number of complete glomeruli mentioned.} If no reference is made to ``complete glomeruli", assume they are complete. \newline
    \textcolor{purple}{- Cores and sections are not relevant to calculations of glomeruli.} \\
    \hline
    number of \textcolor{orange}{globally and locally} sclerosed glomeruli & \textcolor{orange}{Tuple(Numerical, Numerical)} &
    \textcolor{purple}{- Focusing on standard sclerotic lesions for now.} (There are rarer ones like necrotising lesions.) \\
    \hline
    abnormal glomeruli present & Binary &
    \textcolor{purple}{- Only consider non-sclerotic abnormalities.} \newline
    \textcolor{turquoise}{- There are phrases for normal}: normal, no abnormalities, unremarkable. \newline
    \textcolor{turquoise}{- There are phrases for abnormalities}: changes in size and shape, non-sclerotic glomeruli changes. \\
    \hline
    \% chronic change/fibrosis & Numerical \textcolor{orange}{or String-Simple} &
    \textcolor{turquoise}{- What is chronic change?} Chronic change = tubular atrophy + interstitial fibrosis. Can write as 3-tuple: atrophy (y/n), fibrosis (y/n), \% chronic change. \newline
    \textcolor{turquoise}{- What is chronic change not?} Chronic changes in infiltrate are related to transplant biopsies and not relevant here. \newline
    \textcolor{myblue}{- Have schema linking adjectives to \%}, e.g. 5-10\% = ``minimal"/``small", 75-100\% = ``florid"/``complete" \\
    \hline
    transplant & \textcolor{orange}{Tuple(Binary, String-Simple)} &
    \textcolor{purple}{- Identify if transplant (True) or native (False) biopsy.} \newline - \textcolor{turquoise}{Phrases for transplant}: any mention of ``rejection" (including ``no rejection") or cell infiltrate (white blood cells) implies it is a transplant. Sclerosis or fibrosis may be mentioned. \newline - \textcolor{turquoise}{Phrases for native}: no mention of infiltrate and talk solely about sclerosis or fibrosis if the patient has an issue. \newline
    - If included, \textcolor{myblue}{report the grade/quantifier word} associated with severity of rejection, otherwise write ``no grade''. \\
    \hline
    (final) diagnosis & \textcolor{orange}{String-Complex} &
    - Report the final diagnosis of the condition, along with any \textcolor{myblue}{descriptive adjectives}. \textcolor{red}{If no other diagnosis is given, report the rejection diagnosis, and if that isn't there, say [missing].} \\
    \hline
    \end{tabular}}
    \caption{Standard rules/principles to think about when creating entity guidelines: \textcolor{orange}{data types can change}, \textcolor{red}{lack of mention}, \textcolor{purple}{scope of entity}, \textcolor{turquoise}{phrases for/against}, \textcolor{myblue}{severity scores}. Full version of guidelines available in the \texttt{guidelines.xlsx} file.}
    \label{tab:entity-guidelines}
\end{table}
\end{landscape}

\subsection{Prompt Engineering and LM-as-a-Judge Development}

\begin{figure}[h]
    \centering
    \includegraphics[width=1.2\linewidth]{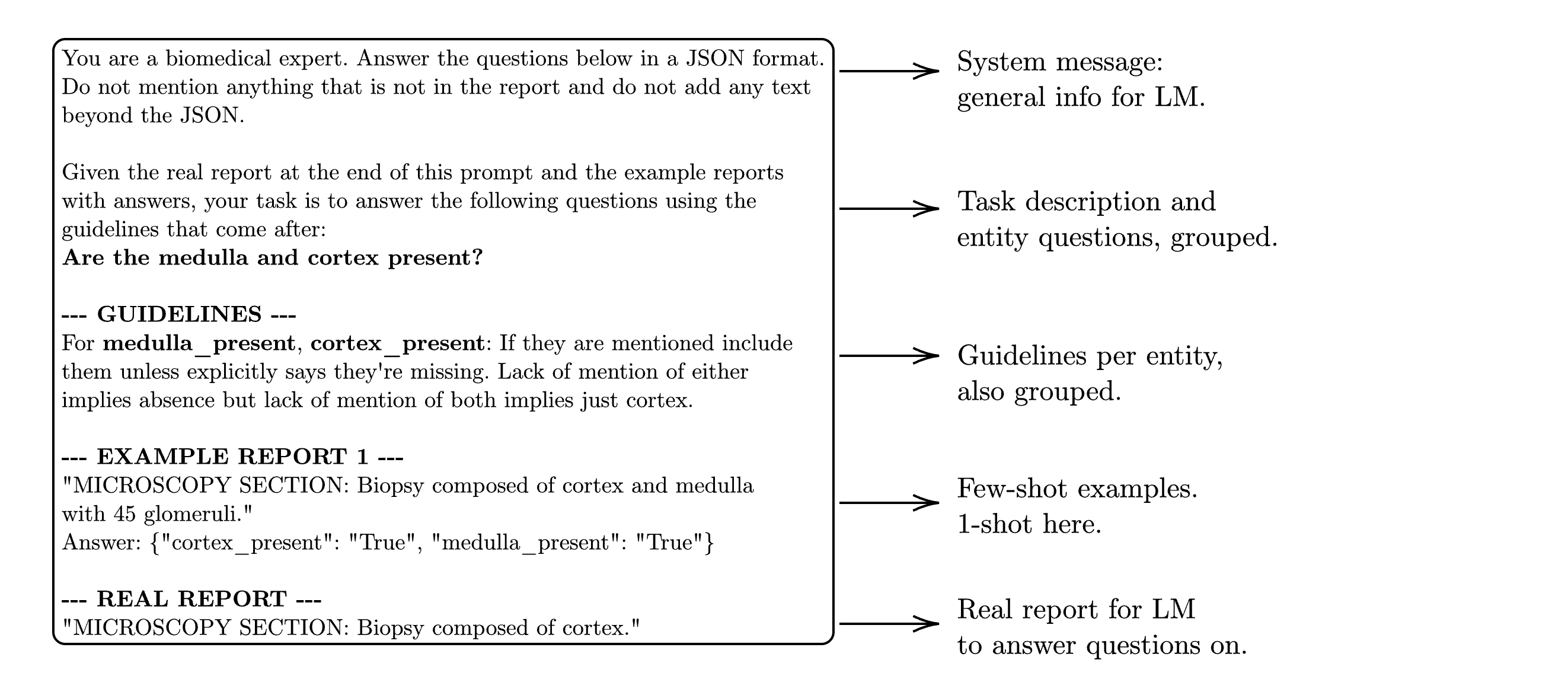}
    \caption{Example entity extraction prompt for two entities.}
    \label{fig:qa-prompt}
\end{figure}

Figure \ref{fig:qa-prompt} illustrates our final prompt structure for two grouped entities (out of the total nine), \textit{cortex\_present} and \textit{medulla\_present}, incorporating system messages (as per other common prompt-based NER methods like GPT-NER \cite{wang2025gptner}), grouped entity questions, grouped entity annotation guidelines, and few-shot examples. 

Our LM-as-a-Judge (LAAJ) evaluation function, shown in Figure \ref{fig:laaj-queries-v2}, underwent iterative refinement to handle string-based entities where exact matching was insufficient. The complete development of this prompt along with an earlier version can be viewed in Supplementary Section 3. The final LAAJ prompt incorporated ``equivalent or similar concepts" language, expert affirmations, answer constraints, and if-else statements for edge cases. Testing across four categories (exact, same concept, similar enough, different) of phrase-pairs (e.g., for ``exact": are phrase A and B exactly the same?) revealed progressive improvement with each prompt iteration (Supplementary Fig. 10), though the final version remained imperfect, incorrectly evaluating approximately 10\% (28/300) of string entities. Despite these limitations, the final LAAJ prompt provided more accurate automated evaluation than previous versions for our specific entity schema.

\begin{figure}[h]
    \centering
    \includegraphics[width=1\linewidth]{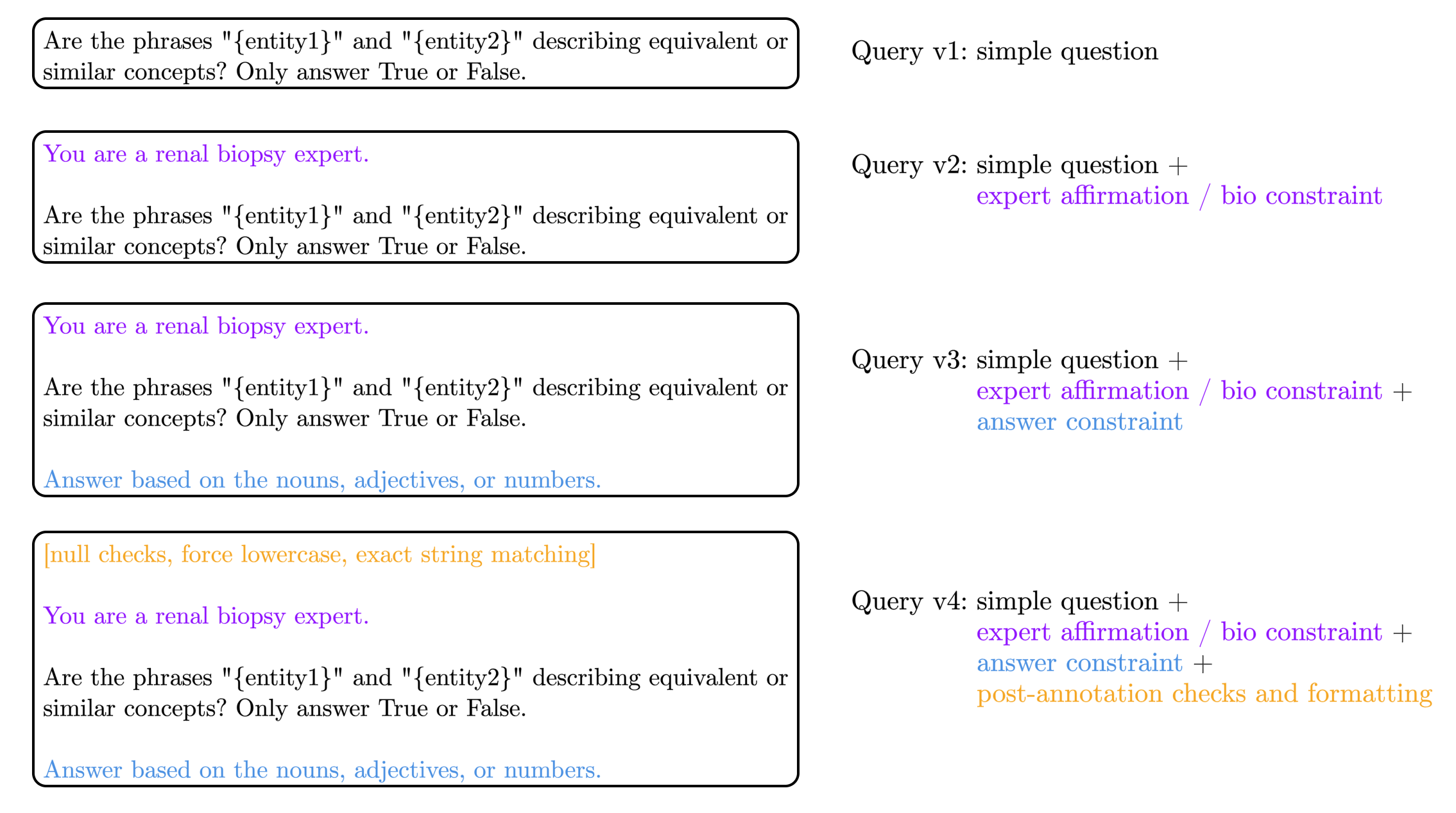}
    \caption{Development of final LM-as-a-Judge (LAAJ) query.}
    \label{fig:laaj-queries-v2}
\end{figure}

\pagebreak

\subsection{Model Performance and Comparison}

Table \ref{tab:model-comparison} demonstrates the impact of prompt engineering across our five SLM models, with all reports containing parsing errors excluded from evaluation and their annotations set to blank values. Guidelines consistently improved performance in zero-shot settings across all models: Qwen-2.5 improved by +9.4\% (67.7\% to 77.1\%), Gemma 2 by +7.4\% (75.9\% to 83.3\%), Llama 3.2 by +18.6\% (45.5\% to 64.1\%), Phi-3.5 Q4 by +7.3\% (74.5\% to 81.8\%), and Phi-3.5 Q8 by +8.1\% (73.6\% to 81.7\%). Few-shot examples improved performance in no-guidelines settings for four of five models: Qwen-2.5 gained +5.7\% (67.7\% to 73.4\%), Gemma 2 increased by +8.4\% (75.9\% to 84.3\%), Llama 3.2 showed substantial improvement of +37.9\% (45.5\% to 83.4\%), and Phi-3.5 Q8 gained +8.7\% (73.6\% to 82.3\%). The exception was Phi-3.5 Q4, which experienced systematic parsing failures that prevented accurate assessment. Unexpectedly, combining both techniques did not provide much additional benefit, either performance-wise, temporally, or with respect to error degradation. Seven configurations achieved our pre-specified performance threshold of $\geq$80\%, all requiring either guidelines or few-shot examples, with Gemma 2 the best-performing model on average.

JSON parsing errors represented the primary barrier to deployment, with error rates varying substantially across models. Gemma 2 and Qwen-2.5 produced minimal errors (1-8 per run), primarily backslash characters interpreted as escape sequences or double quotes within quoted strings, which are easily correctable by replacing backslashes with forward slashes or double quotes with single quotes. In contrast, Phi and Llama models exhibited higher error frequencies with more severe systematic failures. Llama 3.2's zero-shot without guidelines configuration generated 210 bracket closure errors that, whilst systematic, were potentially correctable through automated post-processing. Both Phi-3.5 variants demonstrated a distinct sequential failure mode in configurations with two-shot examples: Q4 and Q8 processed approximately 335 and 322 reports respectively in two-shot with guidelines before undergoing catastrophic degradation and producing incoherent text for all remaining reports. The Q4 variant exhibited similar behaviour in the two-shot without guidelines configuration, generating 114 total errors through the same sequential failure pattern. Whilst the cause of this progressive model degradation during extended processing sessions remains unclear, possible explanations include memory exhaustion on our limited hardware (16GB RAM), KV cache accumulation across sequential inference calls, or quantisation-related instability under extended processing with longer context lengths. These sequential failures make quantised Phi models unsuitable for automated processing of large report batches.

\begin{table}[h]
\fontsize{9}{11}\selectfont
\begin{tabular}{|c||c|c|c|c|}
\hline
\makecell{\textbf{Prompt Format $\rightarrow$} \\ \textbf{Model Name $\downarrow$}}
 & \makecell{\textbf{Zero-shot} \\ \textbf{without guidelines}} & \makecell{\textbf{Two-shot} \\ \textbf{without guidelines}} & \makecell{\textbf{Zero-shot} \\ \textbf{with guidelines}} & \makecell{\textbf{Two-shot} \\ \textbf{with guidelines}} \\
\hline
\hline
\textbf{Qwen-2.5} & 67.7\%$^{\textcolor{red}{\text{*1}}}$ & 73.4\%$^{\textcolor{red}{\text{*8}}}$ & 77.1\%$^{\textcolor{red}{\text{*1}}}$ & 73.3\%$^{\textcolor{red}{\text{*4}}}$ \\
1.5B, FP16 & 30.9s/report & 26.2s/report & 27.4s/report & 26.6s/report \\
\hline
\textbf{Gemma 2} & 75.9\%$^{\textcolor{red}{\text{*1}}}$ & 84.3\%$^{\textcolor{red}{\text{*1}}}$ & 83.3\%$^{\textcolor{red}{\text{*1}}}$ & 84.3\%$^{\textcolor{red}{\text{*2}}}$ \\
2B, FP16 & 35.6s/report & 37.3s/report & 37.9s/report & 36.6s/report \\
\hline
\textbf{Llama 3.2} & 45.5\%$^{\textcolor{red}{\text{*210}}}$ & 83.4\%$^{\textcolor{red}{\text{*1}}}$ & 64.1\%$^{\textcolor{red}{\text{*109}}}$ & 83.4\%$^{\textcolor{red}{\text{*1}}}$ \\
3B, Q8 & 34.8s/report & 32.5s/report & 33.5s/report & 35.2s/report \\
\hline
\textbf{Phi-3.5 Mini} & 74.5\%$^{\textcolor{red}{\text{*11}}}$ & 64.5\%$^{\textcolor{red}{\text{*114}}}$ & 81.8\%$^{\textcolor{red}{\text{*9}}}$ & 74.1\%$^{\textcolor{red}{\text{*65}}}$ \\
3.8B, Q4 & 43.6s/report & 54.0s/report & 44.1s/report & 55.7s/report \\
\hline
\textbf{Phi-3.5 Mini} & 73.6\%$^{\textcolor{red}{\text{*14}}}$ & 82.3\%$^{\textcolor{red}{\text{*3}}}$ & 81.7\%$^{\textcolor{red}{\text{*9}}}$ & 71.4\%$^{\textcolor{red}{\text{*78}}}$ \\
3.8B, Q8 & 51.6s/report & 57.3s/report & 56.1s/report & 72.0s/report \\
\hline
\end{tabular}
\caption{Table comparing accuracy and report iteration times for different models with zero-shot and two-shot prompts over 400 annotated reports. \textcolor{red}{*x} signifies the number of reports with a parsing error when converting the SLM's generated string to a JSON.}
\label{tab:model-comparison}
\end{table}

\begin{comment}
% JSON Parsing Errors Summary by Model
\begin{table}[h]
\centering
\fontsize{9}{11}\selectfont
\begin{tabular}{|l|c|c|l|}
\hline
\textbf{Model} & \textbf{Total Errors} & \textbf{Runs} & \textbf{Most Common Error Types} \\
\hline
\hline
\textbf{qwen2.5:1.5b-instruct-fp16} & 14 & 4 & comments (10), backslash (3), escape (2) \\
\hline
\textbf{gemma2:2b-instruct-fp16} & 5 & 4 & backslash (4), bracketOutsideQuote (1) \\
\hline
\textbf{llama3.2:3b-instruct-q8\_0} & 321 & 4 & noClosingBracket (319), backslash (2) \\
\hline
\textbf{phi3.5:3.8b-mini-instruct-q4\_K\_M} & 199 & 4 & noJsonDict (107), noJson (61), jsonInJson (6) \\
\hline
\textbf{phi3.5:3.8b-mini-instruct-q8\_0} & 104 & 4 & noJson (76), comments (10), noDoubleQuotes (8) \\
\hline
\hline
\textbf{Total} & 643 & 20 & \textbf{Average: 32.2 errors per run} \\
\hline
\end{tabular}
\caption{JSON parsing errors by model. Gemma 2 demonstrates exceptional reliability with minimal errors, while Llama 3.2 and Phi models show significant JSON formatting challenges.}
\label{tab:json-errors-by-model}
\end{table}
\end{comment}

Table \ref{tab:combined-model-comparison} presents comprehensive performance comparison across all (including non-SLM) approaches using optimal configurations for each SLM model. Our SLM workflow achieved the highest accuracy with Gemma 2 reaching 84.3\% when using the two-shot with guidelines configuration, followed closely by Llama 3.2 (83.4\%, two-shot with guidelines) and Phi-3.5 Q8 (82.3\%, two-shot without guidelines). Processing times ranged from 26.2-72.0 seconds per report, enabling complete dataset annotation (n=2,111) within two days on standard NHS laptop hardware. The 10 top-performing SLM configurations ($\geq$75.9\%) outperformed alternative methods: spaCy (74.3\%), BioBERT-SQuAD (62.3\%), RoBERTa-SQuAD (59.7\%), and GLiNER (60.2\%), though these latter methods were substantially faster with spaCy processing the entire 400-report sample in under seven seconds.

Entity-level analysis revealed distinct performance patterns across model types. For simple binary entities like \textit{cortex\_present}, all SLMs achieved $\geq$91\% accuracy, with Gemma 2 leading at 96.0\% - closely matching spaCy (96.8\%) and BioBERT (96.0\%). For numerical entities, SLMs generally outperformed traditional approaches. On \textit{n\_total}, SLMs achieved 94.3-97.3\% compared to spaCy (54.3\%) and BioBERT (86.3\%), with similar patterns for \textit{n\_global} (SLMs: 75.3-92.3\%, spaCy: 68.8\%, RoBERTa: 70.5\%). However, spaCy excelled at \textit{n\_segmental} detection (90.0\% vs 74.8-89.3\% for SLMs). Certain entities proved to be challenging across all models. \textit{abnormal\_glomeruli} accuracy ranged 61.0-74.8\% for SLMs versus 51.5-52.5\% for traditional methods, whilst \textit{chronic\_change} showed variable performance (SLMs: 50.0–79.5\%, alternatives: 45.3–64.5\%). The \textit{transplant} entity showed mixed results, with spaCy achieving the highest accuracy (84.8\%) compared to SLMs (68.5–81.5\%). Notably, \textit{final\_diagnosis} extraction favoured SLMs dramatically (88.5–91.8\%) over traditional approaches (0.2–60.5\%), highlighting the advantage of instruction-following capabilities for complex string entities.

\begin{landscape}
\begin{table}[htbp]
\centering
\fontsize{9}{11}\selectfont
\begin{tabular}{|l||c|c|c|c|c||c|c|c|c|c|}
    \hline
    \makecell{\textbf{Model $\rightarrow$} \\ \textbf{Entity $\downarrow$}} & \textbf{Qwen1.5} & \textbf{Gemma2} & \textbf{Llama3.2} & \textbf{Phi3.5 Q4} & \textbf{Phi3.5 Q8} & \textbf{spaCy} & \textbf{BioBERT} & \textbf{RoBERTa} & \textbf{GLiNER} & \textcolor{orange}{\textbf{NuExtract}} \\
    \hline
    \hline
    \textbf{cortex\_present} & 364/400 & 384/400 & 383/400 & 375/400 & 378/400 & 387/400 & 384/400 & 383/400 & 221/400 & \textcolor{orange}{17/20} \\
    \hline
    \textbf{medulla\_present} & 300/400 & 325/400 & 341/400 & 303/400 & 334/400 & 389/400 & 332/400 & 271/400 & 346/400 & \textcolor{orange}{18/20} \\
    \hline
    \textbf{n\_total} & 385/400 & 379/400 & 389/400 & 377/400 & 378/400 & 217/400 & 345/400 & 295/400 & 342/400 & \textcolor{orange}{19/20} \\
    \hline
    \textbf{n\_segmental} & 339/400 & 329/400 & 299/400 & 336/400 & 357/400 & 360/400 & 160/400 & 253/400 & 340/400 & \textcolor{orange}{17/20} \\
    \hline
    \textbf{n\_global} & 301/400 & 350/400 & 349/400 & 357/400 & 369/400 & 275/400 & 145/400 & 282/400 & 256/400 & \textcolor{orange}{18/20} \\
    \hline
    \textbf{abnormal\_glomeruli} & 244/400 & 299/400 & 291/400 & 295/400 & 280/400 & 206/400 & 207/400 & 209/400 & 210/400 & \textcolor{orange}{14/20} \\
    \hline
    \textbf{chronic\_change} & 200/400 & 318/400 & 292/400 & 208/400 & 219/400 & 255/400 & 258/400 & 237/400 & 181/400 & \textcolor{orange}{15/20} \\
    \hline
    \textbf{transplant} & 274/400 & 295/400 & 294/400 & 326/400 & 285/400 & 339/400 & 168/400 & 143/400 & 249/400 & \textcolor{orange}{13/20} \\
    \hline
    \textbf{final\_diagnosis} & 368/400 & 354/400 & 365/400 & 367/400 & 363/400 & 248/400 & 242/400 & 77/400 & 23/400 & \textcolor{orange}{18/20} \\
    \hline
    \hline
    \multicolumn{11}{|c|}{\textbf{Overall Performance}} \\
    \hline
    \hline
    \textbf{all entities} & 2775/3600 & 3033/3600 & 3003/3600 & 2944/3600 & 2963/3600 & 2676/3600 & 2241/2600 & 2150/3600 & 2168/3600 & \textcolor{orange}{149/180} \\
    \hline
    \textbf{accuracy} & 77.1\% & \textbf{84.3\%} & 83.4\% & 81.8\% & 82.3\% & 74.3\% & 62.3\% & 59.7\% & 60.2\% & \textcolor{orange}{82.8\%} \\
    \hline
    \textbf{speed (s/report)} & 27.4 & 36.6 & 32.5 & 44.1 & 57.3 & 0.02 & 10.86 & 3.70 & 1.53 & \textcolor{orange}{1473} \\
    \hline
\end{tabular}
\caption{Combined Entity Extraction Performance Comparison. Left columns show SLM models with their best-performing configurations. Right columns show alternative approaches. NuExtract results are \textcolor{orange}{highlighted orange} as they are incomplete (only tested on 20/400 reports due to time constraints). BioBERT = BioBERT-SQuAD, RoBERTa = RoBERTa-SQuAD.}
\label{tab:combined-model-comparison}
\end{table}
\end{landscape}

\subsection{Evaluation of the Disagreement Modelling Outputs}

\begin{figure}
    \centering
    \includegraphics[width=1\linewidth]{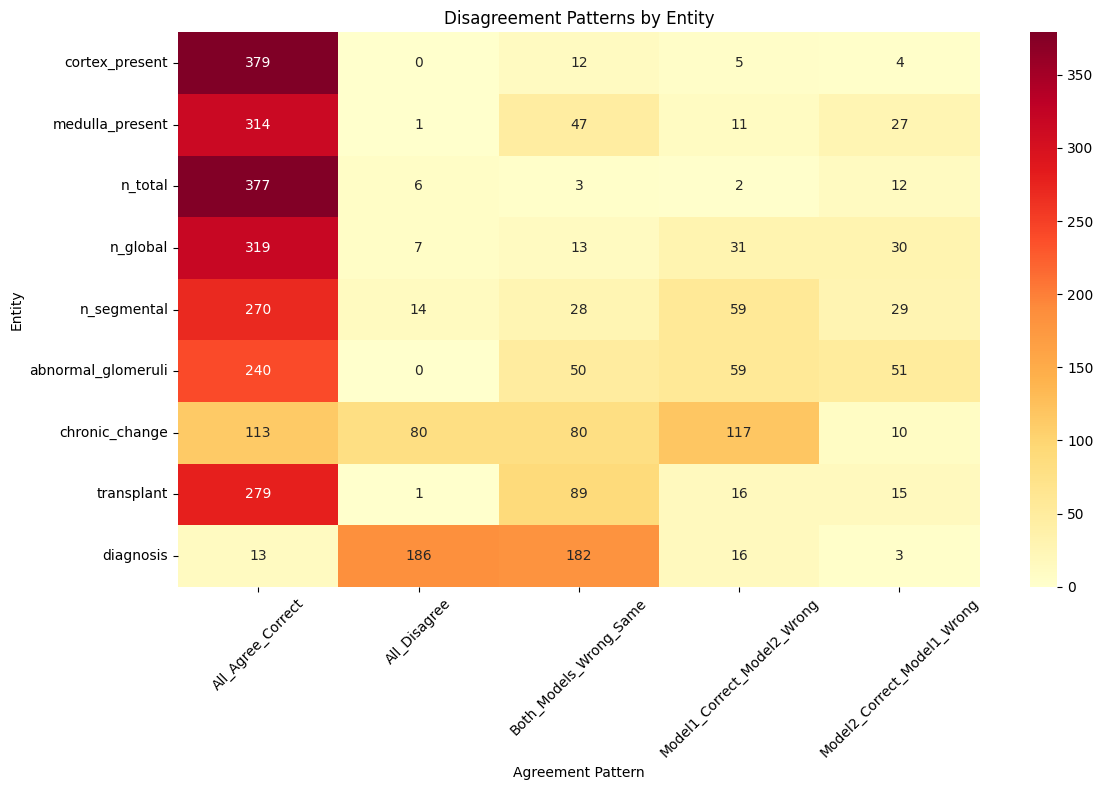}
    \caption{Heatmap showing disagreements between top-performing SLM models (Gemma 2 (Model 1) and Llama 3.2 (Model 2), both with two-shot and guidelines configuration) compared to the ground truth. All\_Agree\_Correct = both predictions match the ground truth, All\_Disagree = both predictions are wrong and differ from each other, Both\_Models\_Wrong\_Same = both predictions are wrong but agree with each other, ModelX\_Correct\_ModelY\_Wrong = model X matches the ground truth but model Y does not.}
    \label{fig:disagreement-heatmap}
\end{figure}

Figure \ref{fig:disagreement-heatmap} shows the three-way comparison between Gemma 2 (two-shot with guidelines) and Llama 3.2 (two-shot with guidelines) against ground truth across 400 reports revealed substantial disagreement patterns that highlight areas requiring clinical review. Overall agreement between all three sources occurred in 64.0\% of cases (2,304/3,600 entity predictions), indicating that nearly two-thirds of predictions achieve consensus. However, systematic disagreements were evident in the remaining cases: both models produced identical incorrect predictions in 14.0\% of cases (504 instances), suggesting consistent misinterpretation of certain clinical features, while complete disagreement between all three sources occurred in 8.2\% of cases (295 instances). 

Model-specific performance analysis revealed differential error patterns, with Gemma 2 demonstrating superior accuracy compared to Llama 3.2. Gemma 2 produced correct predictions when Llama 3.2 failed in 8.8\% of cases (316 instances), while the reverse occurred in only 5.0\% of cases (181 instances). Entity-specific disagreement rates varied dramatically, with \textit{cortex\_present} and\textit{ n\_total} showing minimal disagreement (5.3\% and 5.8\% respectively), moderate disagreement for \textit{medulla\_present} (21.5\%) and \textit{n\_global} (20.3\%), high disagreement for \textit{abnormal\_glomeruli} (40.0\%) and \textit{transplant} (30.3\%), and substantial disagreement for \textit{chronic\_change} (71.8\%) and \textit{final\_diagnosis} (96.8\%).

The disagreement framework's clinical priority assessment identified 21.4\% of all entity predictions (772 cases) as requiring high-priority clinical review, involving high clinical importance entities (\textit{final\_diagnosis}, \textit{transplant}, \textit{n\_total}, \textit{n\_global}, \textit{abnormal\_glomeruli}) where models disagreed. Medium-priority reviews comprised 10.9\% of cases (394 instances), typically involving disagreements on \textit{cortex\_present}, \textit{medulla\_present}, or \textit{chronic\_change} assessments. Low-priority reviews accounted for the remaining 3.6\% (130 instances) for \textit{n\_segmental}. The remaining 64.0\% of cases required no review due to consensus between all sources. Reports with multiple disagreements were prioritised for clinical feedback, with 15/400 reports showing 6–9 disagreements requiring immediate attention. The most problematic cases involved systematic failures where one model produced empty predictions (particularly affecting high-priority entities), suggesting processing errors rather than semantic disagreements. There are also tougher cases of partial matches which need to be assessed on a case-by-case basis, e.g., ground truth = ``no rejection, chronic changes" versus predicted = ``no rejection".

A caveat to these findings: the high disagreement rates for complex string entities (\textit{final\_diagnosis}: 96.8\%, \textit{chronic\_change}: 71.8\%) likely overestimate true model disagreement due to limitations in the LAAJ comparison function. The disagreement framework uses separate LAAJ comparisons for each pairwise evaluation (Model1 vs Ground Truth, Model2 vs Ground Truth, Model1 vs Model2), which can produce inconsistent classifications - two predictions may each be judged equivalent to the ground truth individually, yet be judged as disagreeing with each other, since semantic similarity via SLM comparison is not transitive. The individual evaluation accuracies for these entities are substantially higher (Gemma 2: 79.5\% and 88.5\%; Llama 3.2: 73.0\% and 91.3\%) than implied by the disagreement framework. Calculating each model's accuracy as All\_Agree\_Correct + ModelX\_Correct\_ModelY\_Wrong yields much lower figures: Gemma 2 achieves 57.5\% and 7.3\% for \textit{chronic\_change} and \textit{final\_diagnosis} respectively, whilst Llama 3.2 achieves 30.8\% and 4.0\%. This discrepancy arises from the LAAJ classifying semantically equivalent predictions as disagreements - a conservative behaviour that may be acceptable in clinical settings where missing genuine disagreements is more costly than flagging unnecessary reviews, but which overestimates true model disagreement for these entities.

\section{Discussion}

\subsection{Principal Findings}
We have developed an efficient annotation workflow that produces structured outputs for medical reports, demonstrating that smaller language models ($\leq$5B parameters) can achieve clinically useful accuracy (up to 84.3\%) for structured information extraction from paediatric histopathology reports while operating within typical NHS infrastructure. Our system is designed specifically for constrained clinical environments, operating on CPU infrastructure with sensitive data that cannot leave local servers, successfully achieving efficient (using our slowest SLM and a standard GOSH laptop, it would take $<$2 days to process the total 2,111 reports in our dataset), flexible (adaptive to different annotation schemas) and accurate ($>$80\% accuracy for top-performing models) structured information extraction.

Our evaluation of five SLM models revealed that prompt engineering techniques consistently improve performance across all architectures. We took inspiration from the Banff classification for renal transplant biopsies \cite{loupy2022thirty} to construct simplified GOSH-specific guidelines for both transplant and native biopsies. These guidelines increased accuracy by 7.3–18.6\% across models, whilst few-shot examples provided comparable benefits, increasing accuracy by 5.7–37.9\% in four out of five cases, though a decrease of 10\% was seen for Phi-3.5 Mini Q4. Seven configurations across four models (Gemma 2, Llama 3.2, Phi-3.5 Q4, and Phi-3.5 Q8) achieved our pre-specified threshold of $\geq$80\% accuracy. JSON parsing emerged as a critical deployment consideration, with Gemma 2 and Qwen-2.5 producing minimal errors (1–8 per run) while Phi models produced incoherent responses in the second half of annotation, rendering them unsuitable for large-batch processing. While Bumgardner et al. \cite{bumgardner2024local} found success with quantised LLaMA 13B models for pathology extraction, our results suggest quantisation can introduce instability - smaller full-precision models may be more reliable for automated annotation workflows.

Entity-level analysis revealed distinct performance patterns highlighting the complexity of medical information extraction. All SLMs achieved $\geq$91\% accuracy on simple binary entities like \textit{cortex\_present}, with performance comparable to, though slightly lower than, traditional NLP approaches. However, SLMs demonstrated clear advantages for complex string entities, particularly \textit{final\_diagnosis} extraction (88.5-91.8\% vs 0.2-60.5\% for traditional methods), where few-shots and/or guidelines were able to handle the entity variability. This aligns with broader findings that prompt-based approaches outperform traditional NER for entities with high lexical variability \cite{munnangi2024fly}. Other challenging entities like \textit{abnormal\_glomeruli} (61.0-74.8\%) and \textit{chronic\_change} (50.0-79.5\%) showed variable performance across all approaches, and may require more finegrained entity schemas that split them into smaller entity categories or more detailed entity guidelines that handle all cases.

Our disagreement modelling framework provides a mechanism for comparing model outputs and prioritising reports for clinical review, though with important caveats. The framework identified 64.0\% consensus between both models and ground truth, with the remaining cases flagged for review based on clinical priority weighting. However, the high disagreement rates for \textit{final\_diagnosis} (96.8\%) and \textit{chronic\_change} (71.8\%) reflect limitations in the LAAJ comparison function rather than true model disagreement - qualitative inspection confirmed that many flagged cases contained semantically equivalent predictions. This conservative behaviour may be acceptable in clinical settings where failing to flag genuine disagreements is more costly than flagging cases unnecessarily, but highlights the challenge of automated semantic comparison for complex medical text. Despite these limitations, the framework successfully identifies systematic failure patterns, such as empty predictions, and enables efficient prioritisation of clinician time toward problematic cases.

Through clinician-guided entity guidelines and few-shot samples, our workflow balances performance, efficiency, and clinical integration with minimal clinician time, requiring just three meetings over three months. The disagreement modelling framework provides a way to evaluate prompt robustness across models and forms the automated part of our semi-automated workflow. The top-performing SLM configurations significantly outperform alternative approaches (spaCy: 74.3\%, BioBERT-SQuAD: 62.3\%, RoBERTa-SQuAD: 59.7\%, GLiNER: 60.2\%) while maintaining interpretability through structured guideline development and sufficiently fast processing times of 27.4–57.3 seconds per report. 

Although conducted at GOSH, a well-funded tertiary paediatric centre in London, our workflow's effectiveness on standard 16GB RAM laptops using CPU-only infrastructure suggests it is readily adoptable by other NHS trusts for research purposes. The approach may perform even better on adult populations, which typically exhibit less clinical variability than our paediatric cohort. Our minimal clinician involvement (three meetings over three months) complements more comprehensive human-in-the-loop frameworks like CliniCoCo \cite{gao2024optimising}, offering a lighter-weight alternative for resource-constrained settings where rapid prototyping of annotation workflows is preferred. However, cross-institutional testing will be needed to assess generalisability and understand variations in documentation practices.

\subsection{Comparison with Alternative Approaches}
We initially tried NER annotation which offered some advantages over the QA approach - mainly, it was able capture precise spans consistently for entities such as numbers and words common to simpler entities like \textit{cortex\_present} and \textit{medulla\_present}. However, the span-based focus presents challenges when single spans could correspond to multiple entities. For example, the phrase ``there are 10 glomeruli" with no other reference to glomeruli tells you, with some reasoning, all the answers to our glomeruli questions - there are 10 total, 0 globally sclerosed, 0 segmentally sclerosed, and 0 non-sclerotic abnormalities. The QA framework integrates a flexible entity schema which can be changed via natural language, making it more suitable for non-data scientist annotators.

Baseline approaches using spaCy with rules and regex-based extraction served as our initial starting point. spaCy models achieved 74.3\% overall accuracy across 400 reports, showing strong performance ($>$80\%) for several entities: \textit{cortex\_present} (387/400, 96.8\%), \textit{medulla\_present} (389/400, 97.3\%), and \textit{n\_segmental} (360/400, 90.0\%). However, spaCy struggled with string-based entities and co-referencing, often phrased as ``There are 10 glomeruli, of which 3 have global sclerosis". spaCy also demonstrated critical limitations for complex entities, achieving only 54.3\% accuracy on \textit{n\_total} (217/400) and catastrophically failing on \textit{final\_diagnosis} extraction with just 0.25\% accuracy (1/400). The high spaCy performance on certain entities reflects cases where defaulting to common answers was sufficient - specifically, \textit{n\_segmental} (359/400 are zero), \textit{transplant} (262/400 are True), and \textit{chronic\_change} (180/400 are 0\%).

BioBERT-SQuAD and RoBERTa-SQuAD models showed overall performance of 62.3\% and 59.7\% respectively across the 400-report evaluation set. BioBERT demonstrated superior performance on \textit{final\_diagnosis} entity extraction (242/400, 60.5\%) compared to RoBERTa (77/400, 19.3\%) likely due to its biomedical training, though both substantially underperformed compared to SLMs. GLiNER achieved 60.2\% overall performance, performing comparably to the BERT models given its BioBERT backbone, though, surprisingly, it struggled with \textit{final\_diagnosis} extraction (23/400, 5.8\%). Processing speeds varied but were quicker than all SLMs: spaCy completed the entire 400-report sample in under 7 seconds (0.02 seconds/report), while BioBERT, RoBERTa, and GLiNER required 10.86, 3.70, and 1.53 seconds per report respectively.

The NuExtract model, a fine-tuned version of Microsoft's Phi-3 model specialised for structured JSON extraction, achieved promising performance of 82.8\% accuracy on a limited 20-report subset, comparable to our best SLMs. However, NuExtract proved unsuitable for deployment due to processing times exceeding 24 minutes per report (1,473 seconds) versus our models' 26.2-72.0 seconds, making it impractical for large-scale annotation tasks.

Our top-performing SLM configurations achieved substantially higher accuracy than all alternative approaches: Gemma 2 (84.3\%), Llama 3.2 (83.4\%), and Phi-3.5 Q8 (82.3\%) all exceeded the 80\% clinical utility threshold while maintaining processing speeds under one minute per report. The SLM advantage was particularly pronounced for complex string entities like \textit{final\_diagnosis} extraction, where our models achieved 88.5-91.8\% accuracy compared to 0.2-60.5\% for traditional NLP methods. Despite this, from the alternative methods, spaCy remains valuable as a complementary annotation strategy before or alongside SLM pipelines for certain entities given its speed, interpretability, and relatively high performance.

\subsection{Limitations and Challenges}
Our accuracy metrics were validated on a subset of 400 reports rather than the full dataset of 2,111. This constraint arose from the iterative nature of entity schema development: any change to the schema required re-annotating all previously labelled reports, making comprehensive annotation impractical. After two complete reannotations following meetings with clinicians, we accepted an entity schema that adequately satisfied our design criteria - balancing biological validity, clinical utility, and simplicity - to enable final workflow performance evaluation. While processing times suggest the full dataset could be annotated within two days, performance on the remaining reports was not directly evaluated. Our initial review of 100 reports, with clinician input, suggested the entity schema covered common diagnoses sufficiently for the narrower renal biopsy domain, but this is not guaranteed to hold across all 2,111 reports.

Despite good overall performance, certain entity types (\textit{chronic\_change}, \textit{abnormal\_glomeruli}) remain challenging and often require reasoning across entire sections or even across multiple report sections. A key limitation of the current workflow is that the LM-as-a-Judge evaluation method, while practical, occasionally disagreed with human assessment (approximately 10\% of string entity comparisons in individual evaluation). This discrepancy was amplified in disagreement modelling, where separate pairwise LAAJ comparisons produced substantially lower accuracy estimates for \textit{final\_diagnosis} and \textit{chronic\_change} - individual evaluation accuracies for these entities were 20–87\% higher than implied by the disagreement framework. While the other seven entities showed no such discrepancy, the imperfect LAAJ function for complex string entities restricts the workflow's use to research settings only.

Our prompt engineering results revealed an unexpected finding: combining few-shot examples with guidelines did not meaningfully improve results across any metric - time, accuracy, or error frequency. This is counterintuitive, as one might expect complementary benefits from both structured guidance and concrete examples. Whilst few-shot prompting is well-studied in the literature \cite{brown2020language, yang2021survey}, guidelines remain less explored despite offering greater interpretability for clinical collaborators and aligning more naturally with established classification systems such as Banff \cite{loupy2022thirty}. Munnangi et al. \cite{munnangi2024fly} found that definition augmentation alone improved biomedical NER performance, which supports our finding that guidelines can be effective independently. Whether an optimal balance exists between these approaches, or whether they should be treated as alternatives rather than complements, remains an open question. Prompt sensitivity also poses a challenge: small changes in guidelines or prompt structure should not significantly alter results, yet this stability is difficult to guarantee. This is particularly important in human-in-the-loop workflows where prompt changes are made rapidly, often without explicit version control.

\subsection{Future Work}
Future work should focus on creating more granular entity schemas covering glomerular abnormalities, cell infiltrate composition, vascular dysfunction, and standardised diagnostic terminology, particularly separating rejection grades from diseases rather than combining them into a single \textit{final\_diagnosis} string. An effective approach could involve categorising entities into healthy patients (no sclerosed glomeruli), standard diseased patients (globally sclerosed glomeruli), complex diseased patients (multiple abnormality types), and edge cases, with the aim of aligning human annotations with SLM outputs. Critical research questions for the workflow include optimising clinical touchpoint strategies beyond our three meetings over three months and determining the minimum number of reports requiring qualitative evaluation to establish robust guideline schemas before performance deteriorates.

Performance improvements to cheaper models could be achieved through hybrid approaches: combining spaCy outputs with Gemma 2's performance on the \textit{abnormal\_glomeruli}, \textit{chronic\_change} and \textit{final\_diagnosis} entities would achieve 81.6\% - comparable performance to our top-performing SLMs and quicker annotation enabled by partial spaCy usage. More generally, these hybrid approaches can balance the trade-offs of different models - spaCy for binary entities and simpler numerical entities, alongside SLMs for co-referenced entities and complex strings. Additionally, ensemble approaches that combine judgements from multiple smaller LMs (e.g., JudgeBlender \cite{rahmani2025judgeblender}) could potentially improve reliability whilst maintaining computational efficiency. Tighter integration between data scientists, clinicians, and SLMs requires improved tooling beyond our current Streamlit interface, potentially as downloadable executables or secure web apps with integrated feedback mechanisms. Studies evaluating AI-integrated annotation workflows across different clinical experience levels, similar to recent automated clinical coding work \cite{gao2024optimising}, could inform optimal human-AI collaboration strategies.

Extending to other biomedical areas like foetal post-mortem analyses or genomics presents opportunities but requires careful scope definition. Our approach benefited from the well-established Banff classification for renal transplant biopsies; similarly, other domains with existing standardised guidelines may be more amenable to this workflow. While renal biopsies offer clear treatment pathways, foetal post-mortems serve multiple purposes: immediate parental closure, future pregnancy planning, and reducing stigma around difficult topics \cite{miscarriage2024gosh, shelmerdine2025radiology}. Success requires identifying clinically-relevant scope a priori - for example, instead of genomics broadly, genomics \textit{for monogenic diseases} might represent a more standardised and thus tractable pathway.

\section{Conclusion}

We have presented an iterative approach to annotating and extracting structured information from histopathology reports using SLMs, incorporating expert knowledge through three key touchpoints for refining annotation guidelines and few-shot examples. We achieve strong annotation performance on a 400-report evaluation sample from a total dataset of 2,111 paediatric renal biopsy reports, outperforming other zero-shot information extraction approaches like spaCy, BERT question-answering models, and GLiNER. Notably, combining guidelines with few-shot examples did not yield additional benefits over using either approach alone, suggesting they may function as alternative rather than complementary strategies. Our workflow includes a disagreement modelling stage for initial automated annotation and prioritisation of reports requiring clinical review. A key practical advantage of our approach is its accessibility - the entire workflow can run on a standard 16GB RAM laptop, making it feasible for research purposes across NHS trusts.

\section{Acknowledgements}

AV completed this work as part of the NHS England Data Science PhD Internship between July and December 2024, in collaboration with Great Ormond Street Hospital. AV would like to thank Jenny Chim, Scarlett Kynoch, the rest of the NHS England Data Science team, Joram Posma, and Kristen Severson for helpful discussions around lightweight training of models, consistent output generation for evaluation, and optimising prompts.

\section{Funding}

JSK, SS, JS, AE, DK, JB, SP, PR, and NS receive funding from the Great Ormond Street Children's Charity and Great Ormond Street Hospital NIHR Biomedical Research Centre. AV, WP, JP, DS, and JH receive funding from NHS England. AV is also supported by UK Research and Innovation [UKRI Centre for Doctoral Training in AI for Healthcare grant number EP/S023283/1]. This article presents independent research funded by the NIHR, and the views expressed are those of the author(s) and not necessarily those of NHS England, NIHR or the Department of Health and Social Care. 

\section{Author Information}

\subsection{Contributions}

All authors were involved in review and editing of the manuscript. AV: Conceptualisation, Methodology, Data Processing, Software, Formal Analysis, Investigation, Writing. JSK: Methodology, Data Extraction, Machine Learning Engineering. SS: Methodology, Machine Learning Engineering. JS: Methodology. WP: Methodology. AE: Data Access, Research Infrastructure. DK: Data Access, Research Infrastructure. JB: Data Extraction. SP: Supervision, Project Support. JP: Supervision, Project Support. DS: Supervision, Conceptualisation, Methodology. JH: Supervision, Methodology. PR: Supervision, Conceptualisation, Methodology, Investigation, Data Extraction, Code Refactoring. NS: Supervision, Project Support, Clinical Interpretation.

\subsection{Corresponding Authors}
Avish Vijayaraghavan (avish.vijayaraghavan17@imperial.ac.uk) and Pavithra Rajendran (pavithra.rajendran@gosh.nhs.uk).

\subsection{Competing Interests}

The authors declare no competing financial or non-financial interests.

\section{Data and Code Availability}

Due to data privacy and institutional governance restrictions, the dataset used in this study is not publicly available. However, the code used for data preprocessing, model development, inference, and evaluation is available on GitHub at \url{https://github.com/gosh-dre/nlp_renal_biopsy}.

%**********************************************%

%\bibliographystyle{naturemag}  
%\bibliographystyle{unsrt}
%\bibliography{updated}
\bibliography{sn-bibliography}

% \bibliographystyle{unsrt}

% \newpage
% \clearpage

% other things you can do: stack pictures, sideways pictures


\newpage
\setcounter{section}{0}
\setcounter{figure}{0}
\setcounter{table}{0}

\begin{center}
{\LARGE\textbf{Supplementary Information}}\\[0.5em]
\end{center}

\vspace{1em}

\section{Annotation and Code Design}

\subsection{Annotation Processes}

\begin{figure}[htbp]
    \centering
    \includegraphics[width=0.95\linewidth]{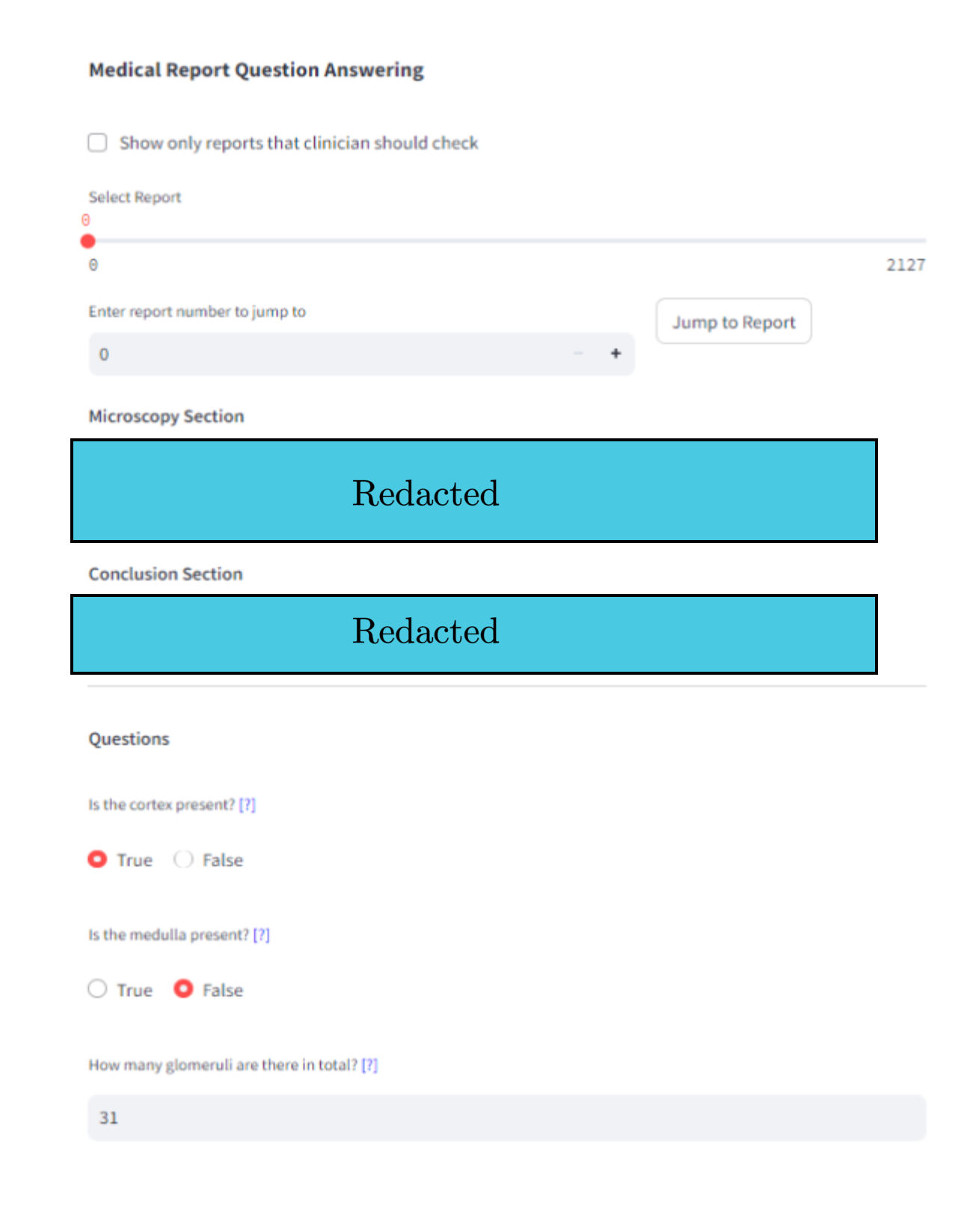}
    \caption{Streamlit app used for QA annotation of reports, displaying three of the nine total entities: \textit{cortex presence}, \textit{medulla presence}, and \textit{number of total glomeruli}. Sensitive patient information in the report has been redacted.}
    \label{fig:annotation-app}
\end{figure}

\textbf{Human annotation via Streamlit annotation app.} We have built a Streamlit web app for QA annotation that can be customised to display relevant sections of biomedical reports. The implementation is discussed in Section \ref{code-design} and part of the app is shown in Figure \ref{fig:annotation-app}. Here, we note down tips that make the annotation process easier. When doing your annotations, it is tempting to start using abbreviations for recurrent phrases (e.g. ``chronic allograft nephropathy" as ``CAN"). It is better to avoid this as it makes the final comparison between your annotations and the SLM smoother. Once you have 10 reports (of the overall recommended 100), we recommend running your SLM model and seeing which entities it does and does not pick up. From this, you can more granularly refine the entity schema and prompts to simultaneously learn the best way to phrase annotations so they are intelligible to humans and produce appropriate outputs from the SLM. This app was inspired by a similar app used for token-classification-based NER annotation, previously developed by Will Poulett.

\textbf{Automated annotation via disagreement modelling.} In order to complete the automated annotation, we run two separate SLM models to get two annotated JSON files. We compare these files for matches and mismatches - any reports with mismatches above a certain threshold are flagged for clinician review via an additional entity in the JSON called \textit{clinician\_check}. The clinician can then check a box at the top of the Streamlit annotation app to view all these reports or view predictions from both models using the Streamlit comparison app described below. The disagreement threshold should be based on number of entities - if you have 2 entities, a threshold of 0.5 (1 entity wrong) is probably reasonable, whereas a lower threshold is suitable for a schema with more entities. For our schema, we notice 3 entities out of 9 are responsible for the majority of errors and so pick a threshold of $0.32$.

\begin{figure}[htbp]
    \centering
    \includegraphics[width=0.95\linewidth]{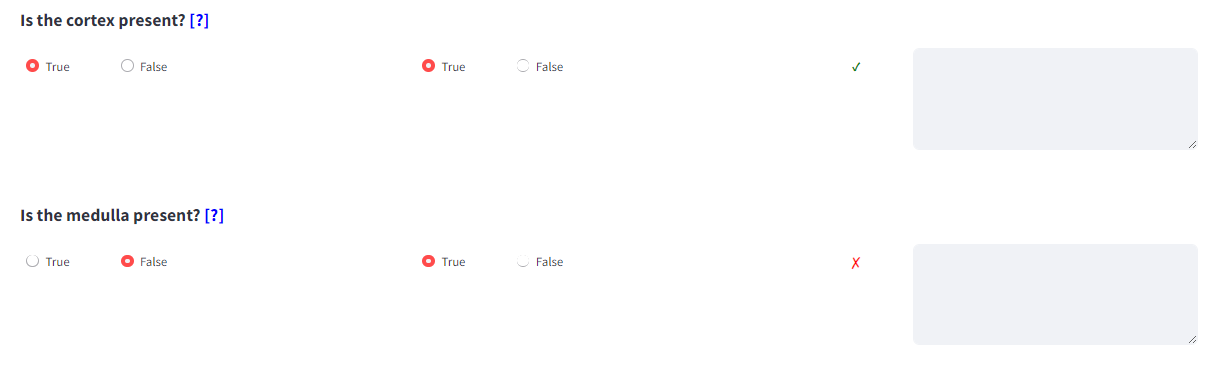}
    \caption{Example report in Streamlit comparison app showing the predictions of two models for \textit{cortex presence} and \textit{medulla presence} entities along with the comment box for qualitative error analysis.}
    \label{fig:comparison-app}
\end{figure}

\textbf{Automated annotation comparison via Streamlit comparison app.} We have designed another Streamlit app to display the predicted results from the automated annotation and add comments on mismatched entities, as shown in Figure \ref{fig:comparison-app}. The app shows the predictions of both models side-by-side along with scores corresponding to their matches (note: these matches are based on our evaluation methods described in Section 3.4 (Experimental Settings) in the main paper), along with a box to add comments per entity per report, with the option to save these comments for further analysis. Comments should include what the correct answer was and which model predicted wrong - doing this over multiple reports can help build general reasons for errors that can feed back into guideline and prompt refinement.

\FloatBarrier

\subsection{Code Design and Implementation} \label{code-design}

The full codebase can be found at \url{https://github.com/gosh-dre/nlp_renal_biopsy}. Our software maps the annotation workflow into three modular code components, centred around a project-specific guidelines spreadsheet, called \texttt{guidelines.xlsx}, shown in Table \ref{tab:entity-guidelines}. This file defines medical entities/phrases to be found in the reports, their codes, their annotation guidelines, and resulting prompts in a format accessible to non-computational researchers. The three distinct component types balance standardisation with flexibility. Utility components handle common tasks such as processing the \texttt{guidelines.xlsx} file and evaluation. Shared components, implemented as base classes, provide the fundamental architecture for question-answering and annotation workflows. Project-specific components, including preprocessing routines, few-shot examples, and report- and domain-specific text processing logic.

The shared components include: (1) a \texttt{qa.py} file which contains the general system message and can be instantiated to implement project-specific question-answering logic with \texttt{Ollama} models, and, (2) a Streamlit-based application \texttt{qa\_app.py} which contains the general app logic for report annotation and can be instantiated to display project-specific reports appropriately. The \texttt{guidelines.xlsx} interfaces with these to produce appropriate prompts for your model and display the report text with space for entity annotations in the Streamlit app. Full steps to apply this annotation workflow to your dataset (along with necessary environment package downloads) can be found in the project repository's \texttt{README.md}.

\begin{table*}
    \centering
    \hspace{-1cm}
    \renewcommand{\arraystretch}{1.4}
    {\scriptsize
    \begin{tabular}{|p{0.14\textwidth}|p{0.18\textwidth}|p{0.18\textwidth}|p{0.08\textwidth}|p{0.19\textwidth}|p{0.15\textwidth}|}
    \hline
    \textbf{Question To Ask} & \textbf{Entity Type} & \textbf{Entity Guidelines} & \textbf{Entity Code} & \textbf{Combined Prompt Question} & \textbf{Combined Guidelines} \\
    \hline
    \textit{Initial questions you get from clinicians and refine from reading reports. This is displayed in Streamlit app.} & \textit{Types in order of complexity: boolean, categorical, numerical, string-simple, string-complex. Can also be combinations e.g. numerical/string-simple.} & \textit{Initial guidelines per entity which you get from reading the reports. Note: use " not ' to avoid Streamlit markdown issues.} & \textit{How entity shown in JSON} & \textit{Certain entities might make sense to group together in final prompt. Add combined questions in here and merge cells.} & \textit{Shortened guidelines that link individual entity guidelines together for the combined prompt questions.} \\
    \hline
    Is the cortex present? & binary & Cortex is the inside of the kidney. If neither medulla or cortex is mentioned, assume the cortex is there and answer True. & cortex\_ present & \multirow{2}{*}{\parbox{0.19\textwidth}{Are the medulla and cortex present?}} & \multirow{2}{*}{\parbox{0.15\textwidth}{If they are mentioned include them unless explicitly says they're missing. Lack of mention of either implies absence but lack of mention of both implies just cortex.}} \\
    \hhline{|----||~|~|}
    Is the medulla present? & binary & Medulla is outside of kidney. (The medulla is covered by the capsule.) If it is not mentioned, assume it is not there and answer False. & medulla\_ present & & \\
    \hline
    \end{tabular}
    }
    \caption{Example of first few rows of \texttt{guidelines.xlsx} file showing the questions that practitioners fill out - first row is the column type, the second row contains descriptions of the column type, and all following rows correspond to our entity information.}
    \label{tab:entity-guidelines}
\end{table*}

\pagebreak

\section{Additional Figures for Exploratory Data Analysis}

\begin{figure}[H]
    \centering
    \includegraphics[width=0.6\linewidth]{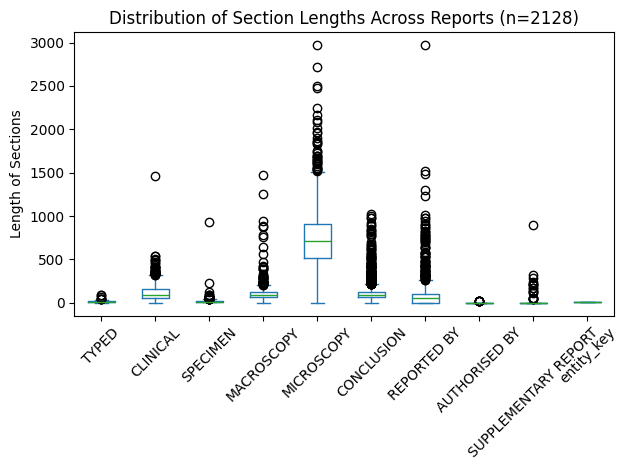}
    \caption{Boxplot showing lengths of each section of the report.}
    \label{fig:section-lengths}
\end{figure}

\begin{figure}[htbp]
    \centering
    \includegraphics[width=0.72\linewidth]{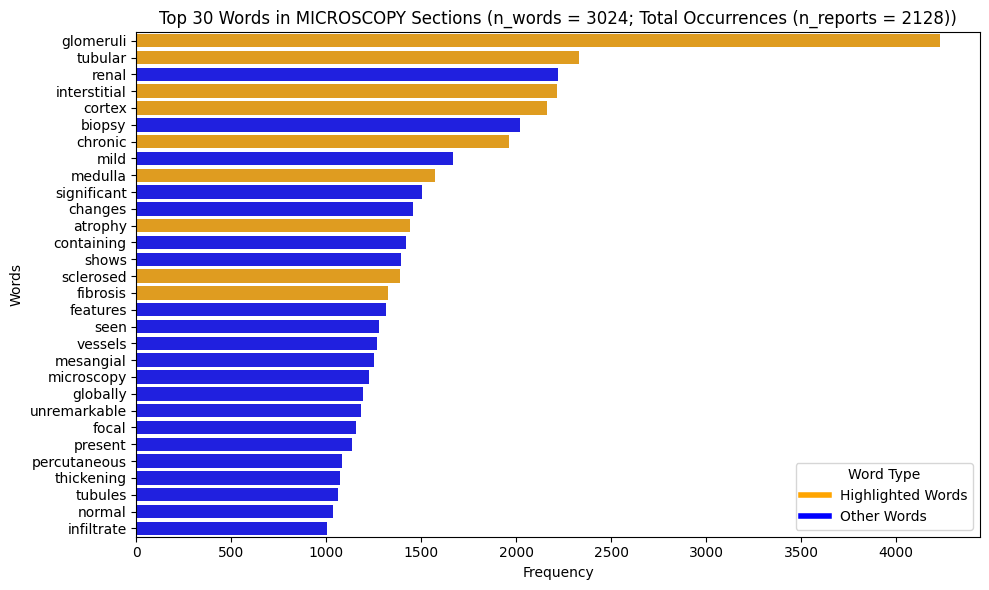}
    \caption{Top 30 words in ``MICROSCOPY" section across reports.}
    \label{fig:wordfreq}
\end{figure}

\begin{figure}[htbp]
    \centering
    \begin{subfigure}[b]{\linewidth}
        \centering
        \includegraphics[width=0.85\linewidth]{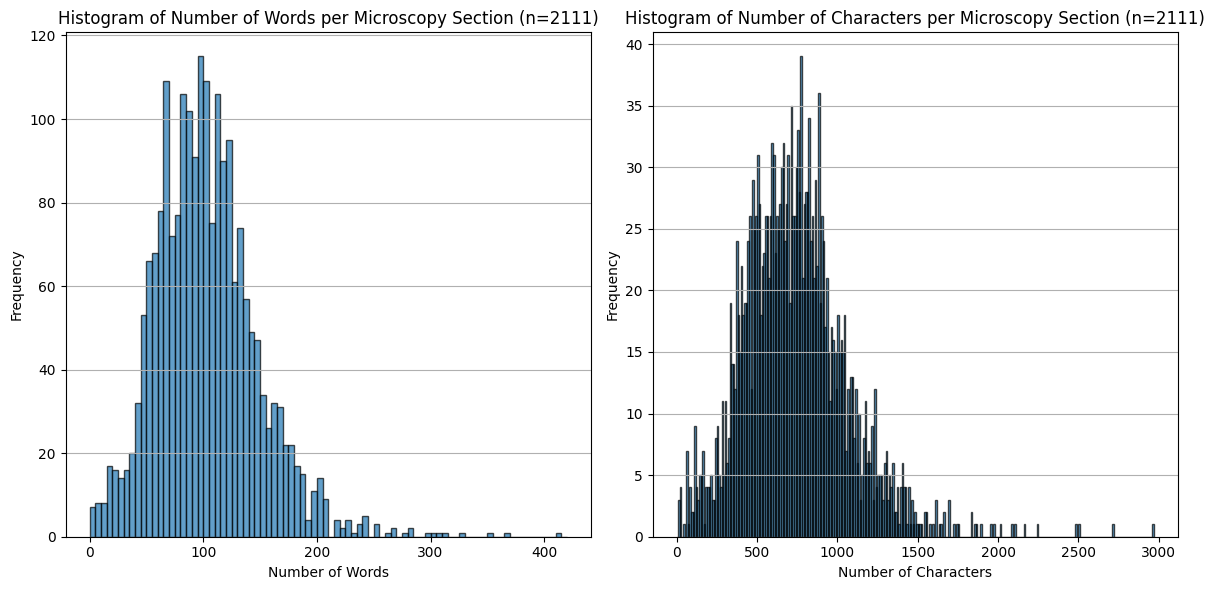}
        \label{fig:hist-micr}
    \end{subfigure}
    
    \begin{subfigure}[b]{\linewidth}
        \centering
        \includegraphics[width=0.85\linewidth]{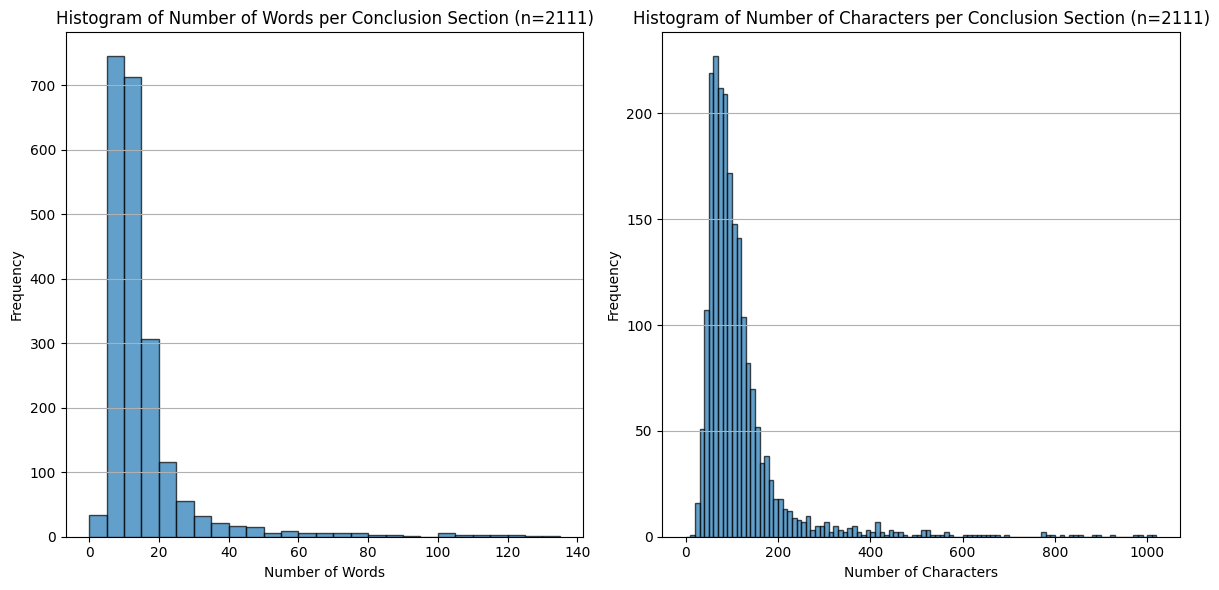}
        \label{fig:hist-conc}
    \end{subfigure}
    
    \caption{Histograms of number of words (left) and characters (right) for the microscopy (top) and conclusion (bottom) sections. There are fewer reports here than in Figures \ref{fig:section-lengths} and \ref{fig:wordfreq} since not all reports with ``MICROSCOPY" sections had ``CONCLUSION" sections.}
    \label{fig:combined-hist}
\end{figure}

The raw renal biopsy dataset consisted of 2,462 reports. The initial segmentation procedure yielded 2,128 reports, of which 2,111 contained both ``MICROSCOPY" and ``CONCLUSION" sections that were required as input to our SLM models. Figure \ref{fig:wordfreq} illustrates the frequency distribution of terms (including orthographic variations identified through Levenshtein distance calculations) across all reports after removing common stop words. Orange highlights are shown for words associated with our entities. Analysis of report sections revealed that the "MICROSCOPY" section was consistently the longest (Figure \ref{fig:section-lengths}). Word and character count distributions for the key sections ``MICROSCOPY" and ``CONCLUSION" are presented in Figure \ref{fig:combined-hist}. The reports exhibited a mean length of 854 $\pm$ 358 words, ranging from 50 to 3493 words. This maximum length corresponds to approximately 4656 tokens (calculated using the standard 1.33 words-to-token ratio), establishing the minimum required context window for language model inference.

\pagebreak

\section{LM-as-a-Judge Prompt Development}

We create multiple pairs of output renal biopsy phrases for the following categories: ``exact", ``same concept", ``similar enough", and ``different" categories. ``Exact" pairs are the exactly same phrase (this is a sanity check for the model), ``same concept" have the exact same meaning but are phrased differently (including abbreviations), ``similar enough" are phrases that are similar enough to be considered the same for our purposes (requiring some inference from the model), and ``different" are phrases which are not the same. We include edge cases from running our model over \textit{string} entities from our guidelines. We test the ability for binary entities to be semantically the same but syntactically different, e.g., ``True" vs ``Present" for a entity's presence, and do not allow any margin of error for numbers e.g. ``14" cannot be accepted for ``15".

We test our language model (LM)-as-a-Judge (LAAJ) function (the model used and prompt) via scores for ``symmetry", ``expert agreement", and ``consistency". For ``symmetry", we run our entity pair as input to our prompt with inputs as $(entity_1, entity_2)$ (forward pass) and then $(entity_2, entity_1)$ (backward pass) to confirm the outputs are exactly the same. For ``expert agreement", we check that our forward pass outputs the same boolean value for similarity as the human (``exact", ``synonyms", and ``similar" are \textit{True}, ``different" is \textit{False}), and, for ``consistency", we check both the forward and backward pass output the same boolean for similarity as the human. LM temperature is always set to zero (though this does not mean completely deterministic results) and so only one trial run per entity pair and score is performed. The aim here is not to achieve perfect performance but to figure out cases in which our LM evaluator outputs something different to what we want so these cases can be avoided during annotation or given to a clinician for further checks.

\subsection{Initial Versions}

\textbf{Note: these were not used for final evaluation.}

A version with HuggingFace \texttt{sentence-transformers} library and cosine similarity was tried initially but setting the similarity threshold proved to be too difficult, particularly for very differently-phrased synonyms, similar-looking but contradictory phrases, and negations.

\begin{figure}[h]
    \centering
    \includegraphics[width=0.95\linewidth]{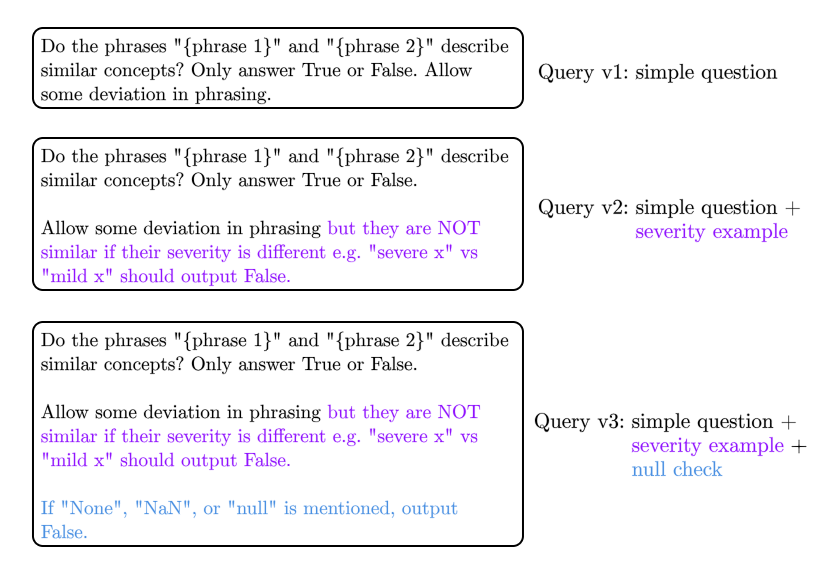}
    \caption{Development of final LM-as-a-Judge (LAAJ) query.}
    \label{fig:laaj-queries}
\end{figure}

\begin{figure}[htbp]
    \centering
    \begin{subfigure}[b]{\linewidth}
        \centering
        \includegraphics[width=0.7\linewidth]{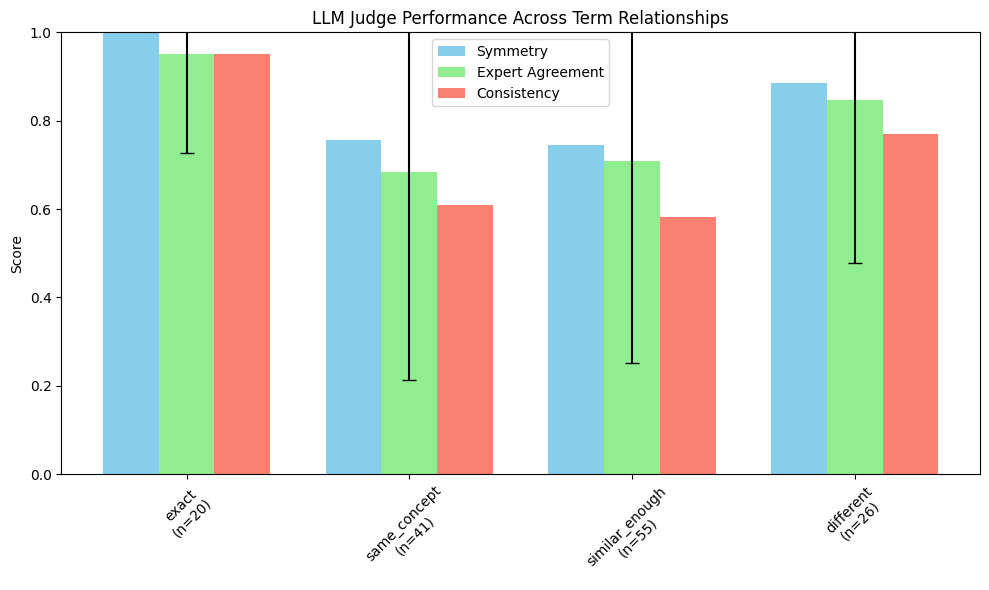}
        \caption{Query v1: simple question}
    \end{subfigure}
    
    \begin{subfigure}[b]{\linewidth}
        \centering
        \includegraphics[width=0.7\linewidth]{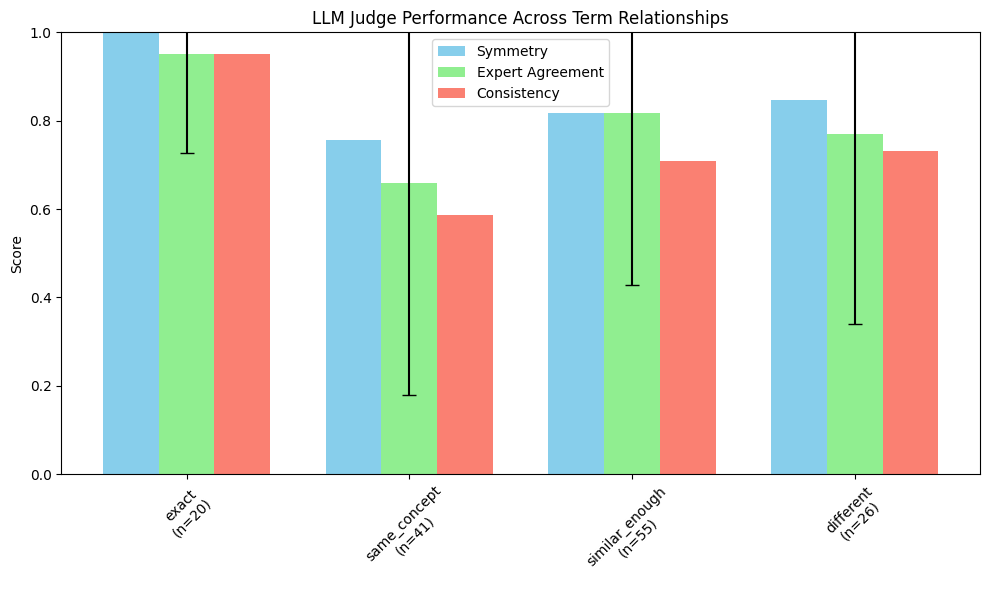}
        \caption{Query v2: simple question + severity example}
    \end{subfigure}
    
    \begin{subfigure}[b]{\linewidth}
        \centering
        \includegraphics[width=0.7\linewidth]{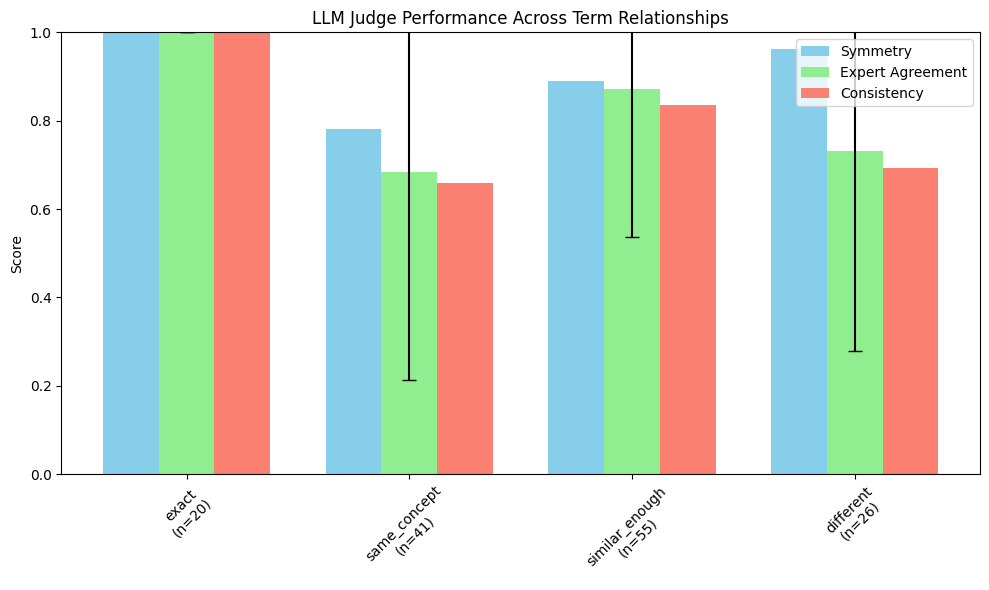}
        \caption{Query v3: simple question + severity example + null check}
    \end{subfigure}
    \caption{Results from Qwen FP16 model across different LAAJ queries. For each plot, from left to right: exact, same\_concept, similar\_enough, different.}
    \label{fig:laaj-qwen-results}
\end{figure}

\begin{figure}[htbp]
    \centering
    \begin{subfigure}[b]{\linewidth}
        \centering
        \includegraphics[width=0.7\linewidth]{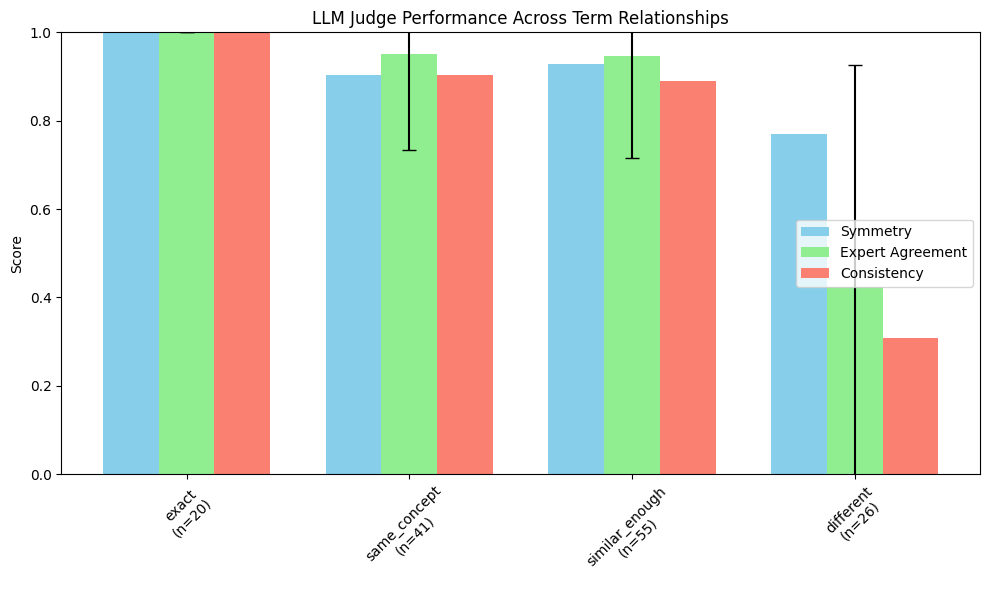}
        \caption{Query v1: simple question}
    \end{subfigure}
    
    \begin{subfigure}[b]{\linewidth}
        \centering
        \includegraphics[width=0.7\linewidth]{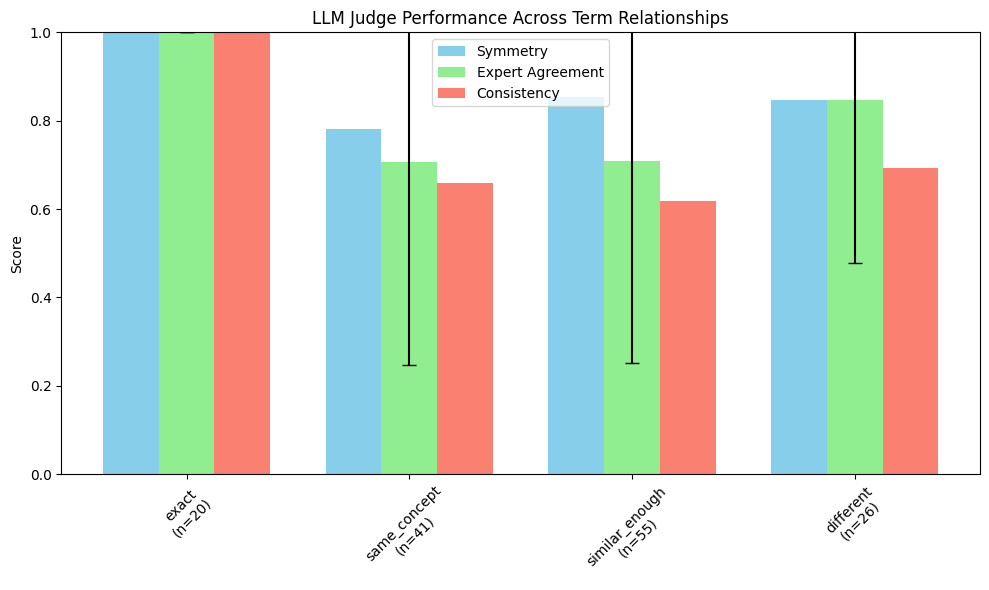}
        \caption{Query v2: simple question + severity example}
    \end{subfigure}
    
    \begin{subfigure}[b]{\linewidth}
        \centering
        \includegraphics[width=0.7\linewidth]{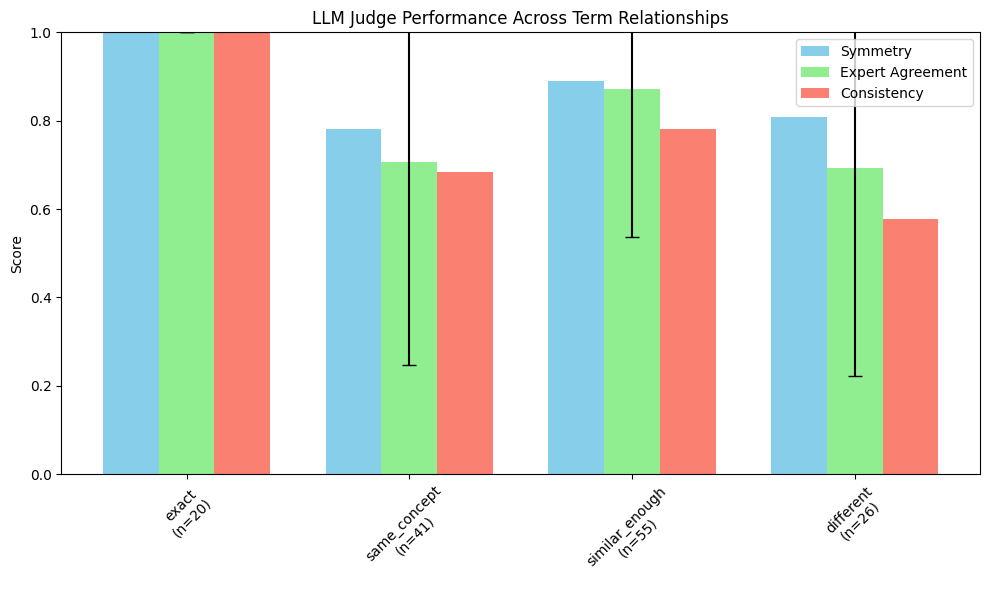}
        \caption{Query v3: simple question + severity example + null check}
    \end{subfigure}
    \caption{Results from Gemma2 2B FP16 model across different LAAJ queries. For each plot, from left to right: exact, same\_concept, similar\_enough, different.}
    \label{fig:laaj-gemma2-results}
\end{figure}

Longer prompts were tried with techniques similar to the input prompt such as few-shots and longer sets of guidelines about different comparisons. However, simplicity seemed to be key to good performance on our examples. We found that a generic example for severity improved performance. We show three slight variations of a simple prompt in Figure \ref{fig:laaj-queries} and their performance using Qwen 2.5 1.5B FP16 in Figure \ref{fig:laaj-qwen-results} and Gemma2 2B FP16 in Figure \ref{fig:laaj-gemma2-results}. We found that Qwen 2.5 1.5B FP16 with Query v3 gave us the best overall performance and this was used for model evaluation in later sections. 

We note that Query v1 for Gemma2 looks good for the first three categories but the poor performance on the ``different" examples made it too unreliable. We tried some of the other versions - at most, we found a decrease in performance of 3\% performance, and so around 30/900 outputs changing, or more accurately, 30/200 given that there were 200 string entities that require LAAJ evaluation. These were mostly nested phrases for the \textit{final\_diagnosis} entity that were poorly predicted by one prompt. The variation in symmetry is surprising and evidence that this workflow is not yet suitable for deployment workflows. Worse still, these prompts seemed to give far too many false positives when evaluating our LM predictions on the real dataset. We tried a different set of prompts shown in the following section and found more accurate downstream performance.

\pagebreak

\subsection{Final Version}

\begin{figure}[h]
    \centering
    \includegraphics[width=1.10\linewidth]{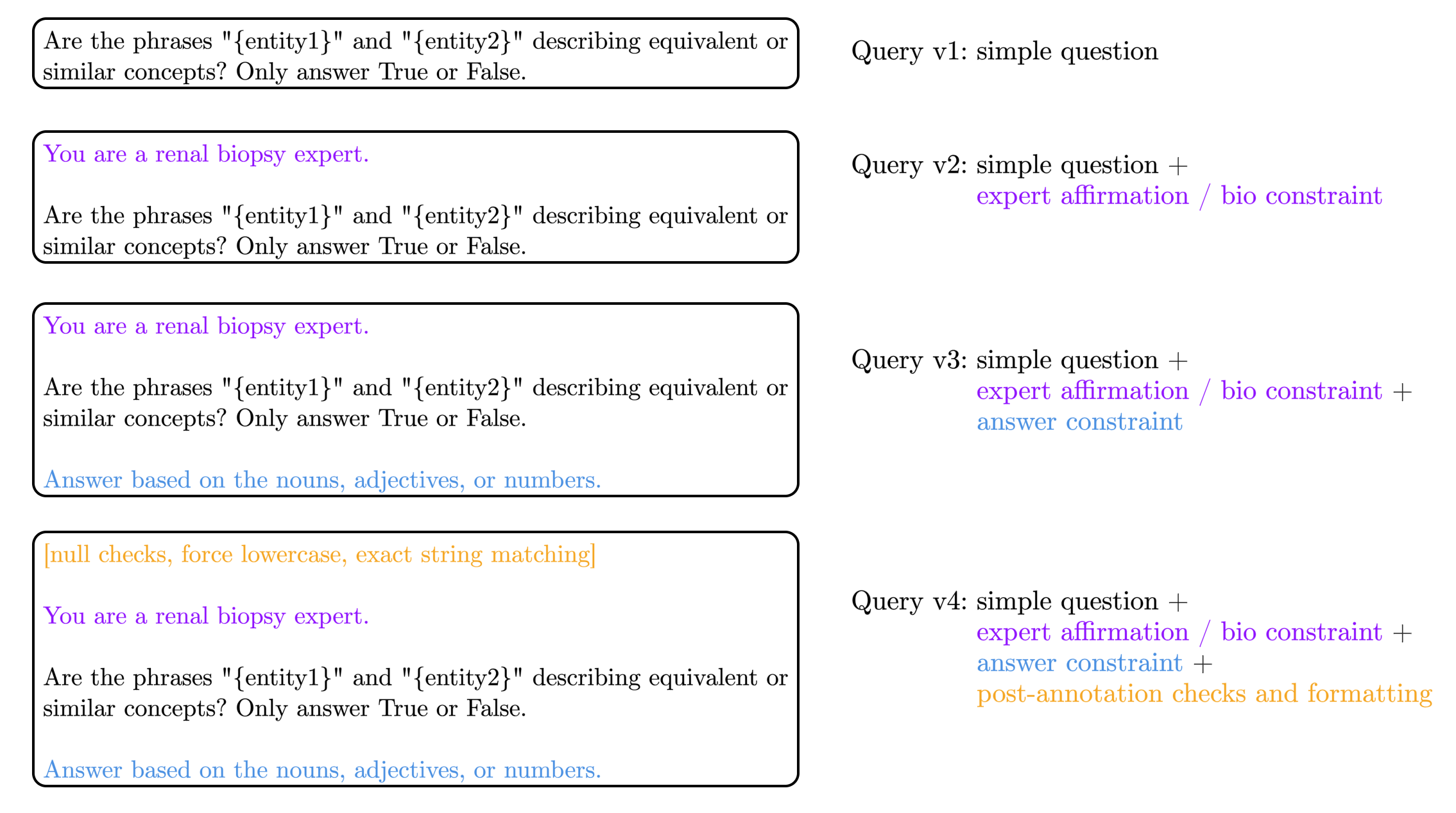}
    \caption{Development of final LM-as-a-Judge (LAAJ) query.}
    \label{fig:laaj-queries-v2}
\end{figure}

One issue with the previous set of prompts is it would either perform well on the \textit{chronic\_change} or \textit{final\_diagnosis} entity but not both. We were able to create our final prompt in Figure \ref{fig:laaj-queries-v2} that balanced the two with reference to ``equivalent or similar concepts", appropriate use of expert affirmations, answer constraints, and if-else statements for edge cases. \textbf{We emphasise that this prompt is not perfect - it does not always evaluate as we would like it to (gets it wrong roughly 10\% of the time on the \textit{string} entities) - and so should be used with care.} 

We show results for this prompt on our test cases with Llama 3.2 3B FP16 in Figure \ref{fig:laaj-llama-results}. The initial simple query was good at spotting differences but poor at all synonyms and extremely poor at exact matches - the symmetry was still high so it appears as though it was defaulting to False for many of these. Inclusion of the expert affirmation made performance much more consistent across the four categories. Inclusion of the answer constraints improved the ``same\_concept" and ``similar\_enough" categories but decreased the performance on ``difference", implying it was defaulting to True. A final set of if-else checks for edge cases was able to bring performance up for the exact matches. The remaining issues for ``difference" include (1) proper handling of ``null/None/NaN" values, (2) subtle rejection differences (Rejection Grade 1A vs 1B, T-cell mediated rejection vs antibody-mediated rejection), (3) the inability to separate different types of sclerosis (global vs segmental, diffuse vs focal), and (4) the occasional confusion of opposite/orthogonal terms (cortex vs medulla, interstitial vs scattered, acute rejection vs chronic rejection). 

The first three issues are understandable, the last indicates much more work is needed for a generalisable medically-suited LAAJ prompt but fortunately, these are not issues that ever happen in actual annotation. In fact, we found that this prompt gave more accurate evaluation than our previous set of queries even though it was slightly worse in overall performance. We assume this is because our set of test pairs cover much broader ground than the actual LM vs human pairs for our two string entities \textit{chronic\_change} and \textit{final\_diagnosis}; in other words, the previous prompt might be better for general renal biopsy schemas but this one is better for our entity schema. Future work could look to build test sets that are more targeted to a specific entity schema.

\begin{figure}[h]
    \centering
    \begin{subfigure}[b]{\linewidth}
        \centering
        \includegraphics[width=0.8\linewidth]{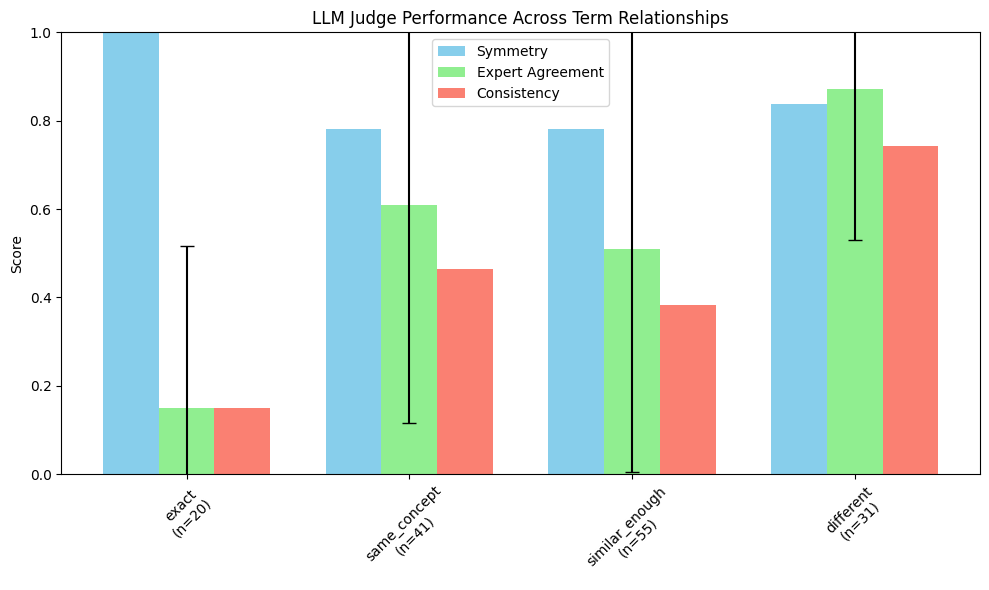}
        \caption{Query v1: simple question}
    \end{subfigure}
    
    \begin{subfigure}[b]{\linewidth}
        \centering
        \includegraphics[width=0.8\linewidth]{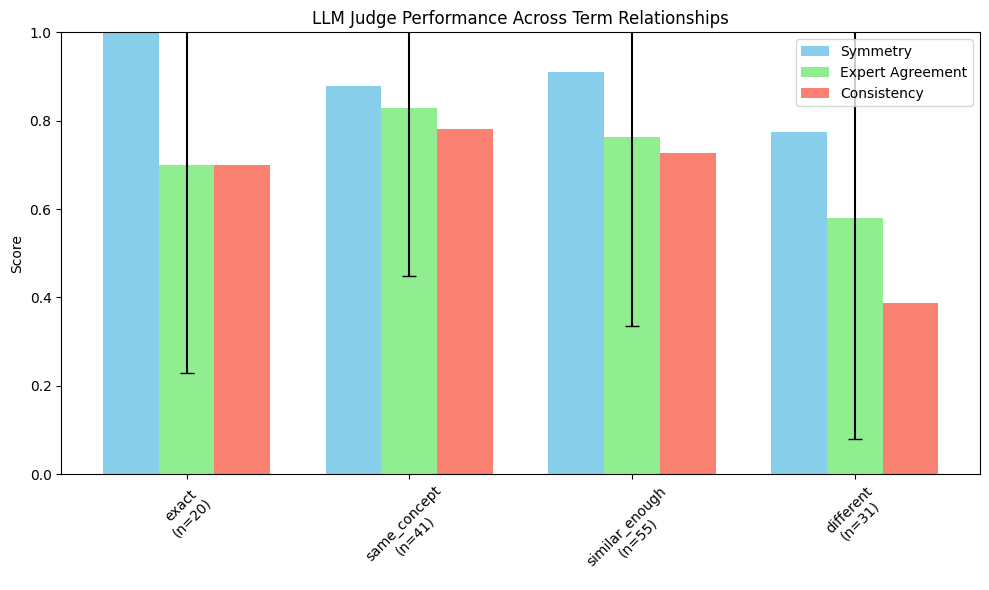}
        \caption{Query v2: simple question + expert affirmation}
    \end{subfigure}
    \caption{Results from Llama 3.2 3B FP16 across different LAAJ queries. For each plot, from left to right: exact, same\_concept, similar\_enough, different. (Continued below...)}
    \label{fig:laaj-llama-results}
\end{figure}

\begin{figure}[p]
    \ContinuedFloat
    \centering
    \begin{subfigure}[b]{\linewidth}
        \centering
        \includegraphics[width=0.8\linewidth]{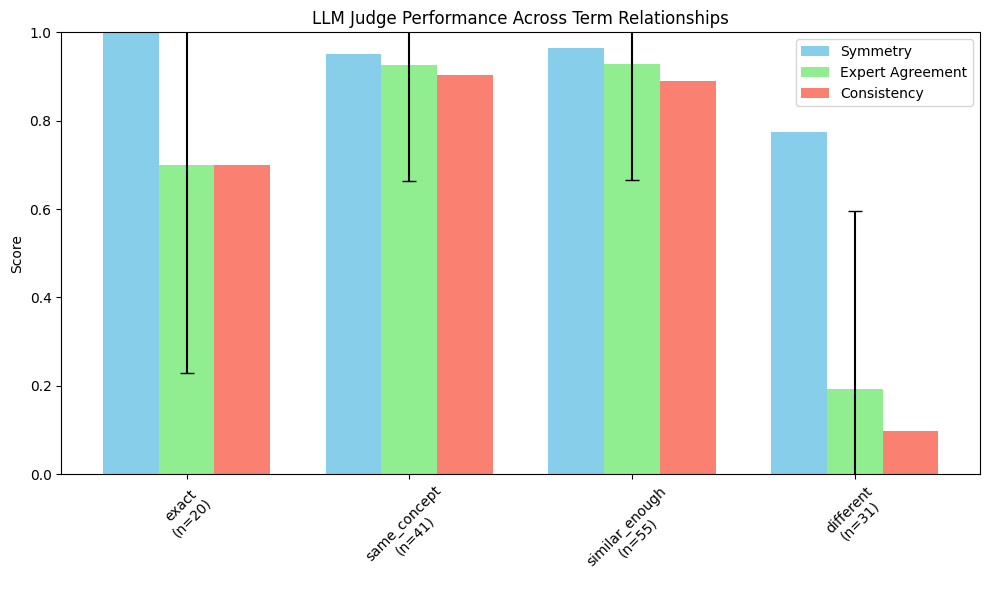}
        \caption{Query v3: simple question + expert affirmation + answer constraints}
    \end{subfigure}
    
    \begin{subfigure}[b]{\linewidth}
        \centering
        \includegraphics[width=0.8\linewidth]{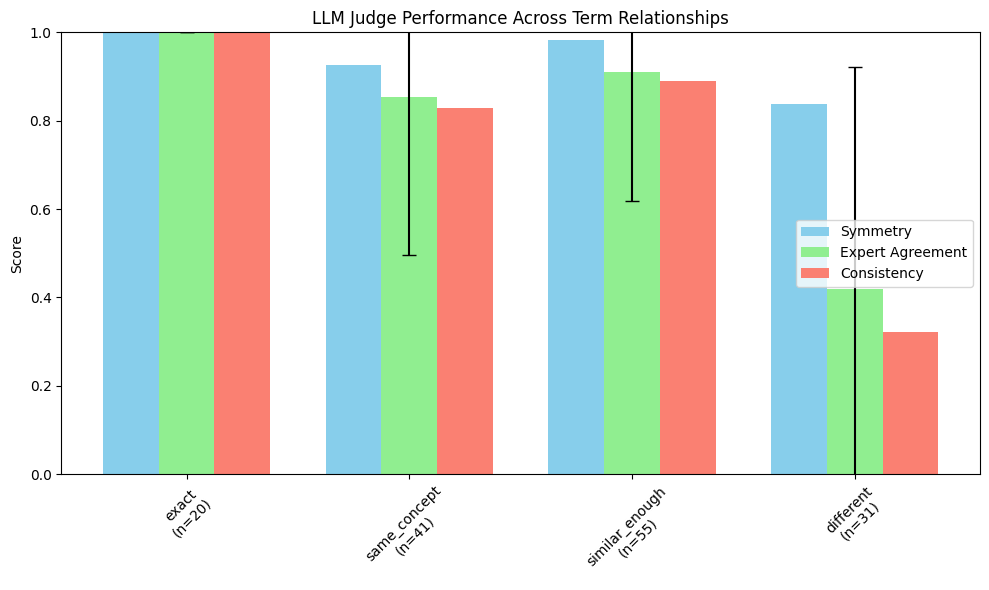}
        \caption{Query v4: simple question + expert affirmation + answer constraints + post-annotation checks}
    \end{subfigure}
    \caption{Continued from above.}
\end{figure}


\end{document}